\documentclass{article}

 \usepackage[preprint]{neurips_2026}

\usepackage[utf8]{inputenc} 
\usepackage[T1]{fontenc}    
\usepackage{hyperref}       
\usepackage{url}            
\usepackage{booktabs}       
\usepackage{amsfonts}       
\usepackage{nicefrac}       
\usepackage{microtype}      
\usepackage{xcolor}         
\usepackage{amsmath}
\usepackage{enumitem}
\usepackage{graphicx}
\usepackage[dvipsnames, svgnames, x11names]{xcolor}
\usepackage{tikz}
\usetikzlibrary{positioning,arrows.meta,calc}
\usepackage{wrapfig} 
\usepackage{multirow}
\usepackage[table]{xcolor}
\usepackage{pifont}
\usepackage{tcolorbox}
\usepackage{cleveref}
\usepackage{pgfplots}
\pgfplotsset{compat=1.18}
\usetikzlibrary{decorations.pathreplacing}
\usepackage{subcaption}
\usepackage{tikz}
\usepackage{hyperref}
\hypersetup{
    colorlinks=true,
    urlcolor=green
}

\title{
    \textcolor{DeepPink}{A}\textcolor{DeepPink}{da}\textcolor{DeepPink}{p}\textcolor{DeepPink}{To}\textcolor{DeepPink}{PA}\textcolor{DeepPink}{S}\textcolor{DeepPink}{S}: 
    \textcolor{DeepPink}{A}mbiguity-aware 
    A\textcolor{DeepPink}{da}ptive 
    S\textcolor{DeepPink}{p}herical 
    \textcolor{DeepPink}{T}ransf\textcolor{DeepPink}{o}rmer for 
    \textcolor{DeepPink}{PA}noramic 
    \textcolor{DeepPink}{S}emantic 
    \textcolor{DeepPink}{S}egmentation
}

\author{%
  Soumyaratna Debnath \\
  EmPACT LAb @ NTU Singapore \\
  \texttt{soumyara004@e.ntu.edu.sg} \\
  \And
  Weiming Zhang \\
  HKUST \\
  \texttt{wzhang915@connect.hkust-gz.edu.cn} \\
  \AND
  Shriram Damodaran \\
  EmPACT LAb @ NTU Singapore \\
  \texttt{shriram003@e.ntu.edu.sg} \\
  \And
  Dingwen Xiao \\
  HKUST \\
  \texttt{dxiaoaf@connect.hkust-gz.edu.cn} \\
  \And
  Addison Lin Wang~\thanks{Corresponding Author} \\
  EmPACT LAb @ NTU Singapore \\
  \texttt{linwang@ntu.edu.sg} \\
}

\definecolor{cvprBlue}{rgb}{0.21,0.49,0.74}
\definecolor{DeepPink}{rgb}{1,0.078,0.576}
\definecolor{GRAY}{rgb}{0.4, 0.4, 0.4}
\hypersetup{
    breaklinks=true,
    colorlinks=true,
    allcolors=DeepPink
}

\newtcolorbox{lightbluebox}{
  colback=blue!5,
  colframe=blue!10,
  boxrule=0.2pt,
  arc=1mm,
  left=1mm,
  right=1mm,
  top=1mm,
  bottom=1mm,
  before skip=4pt,
  after skip=4pt
}

\makeatletter
\renewcommand{\@toptitlebar}{}
\renewcommand{\@bottomtitlebar}{}
\makeatother

\begin{document}

\begin{tikzpicture}[remember picture, overlay]

\node[anchor=north west, inner sep=0pt]
at ($(current page.north west)+(0.72in,-0.43in)$)
{
    \includegraphics[
        width=3.0cm,
        trim=3 3 3 3,
        clip
    ]{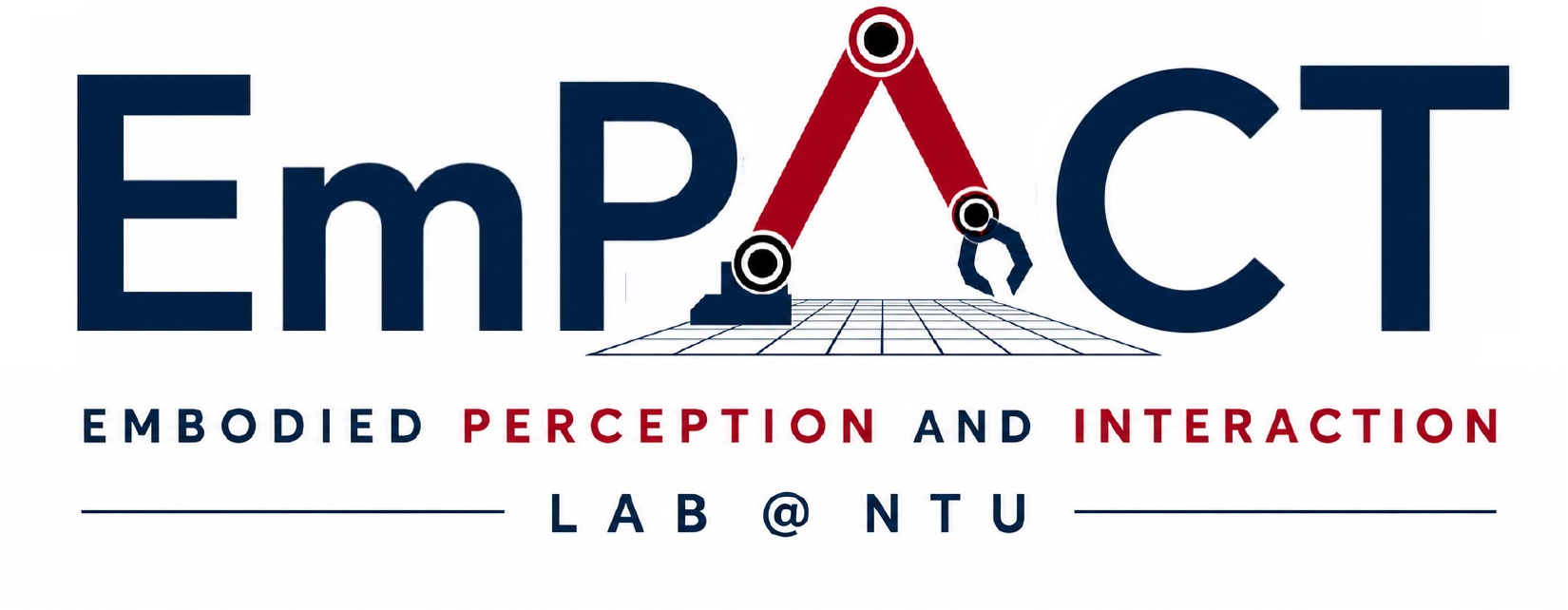}
};

\draw[
    black!20,
    line width=0.35pt
]
($(current page.north west)+(0.72in,-0.98in)$)
--
($(current page.north east)+(-0.72in,-0.98in)$);

\end{tikzpicture}

\maketitle

\begin{figure}[h!]
\vspace{-10pt}
\label{sec:teaser}
    \centering
    \vspace{-10pt}
    \includegraphics[width=1.0\linewidth]{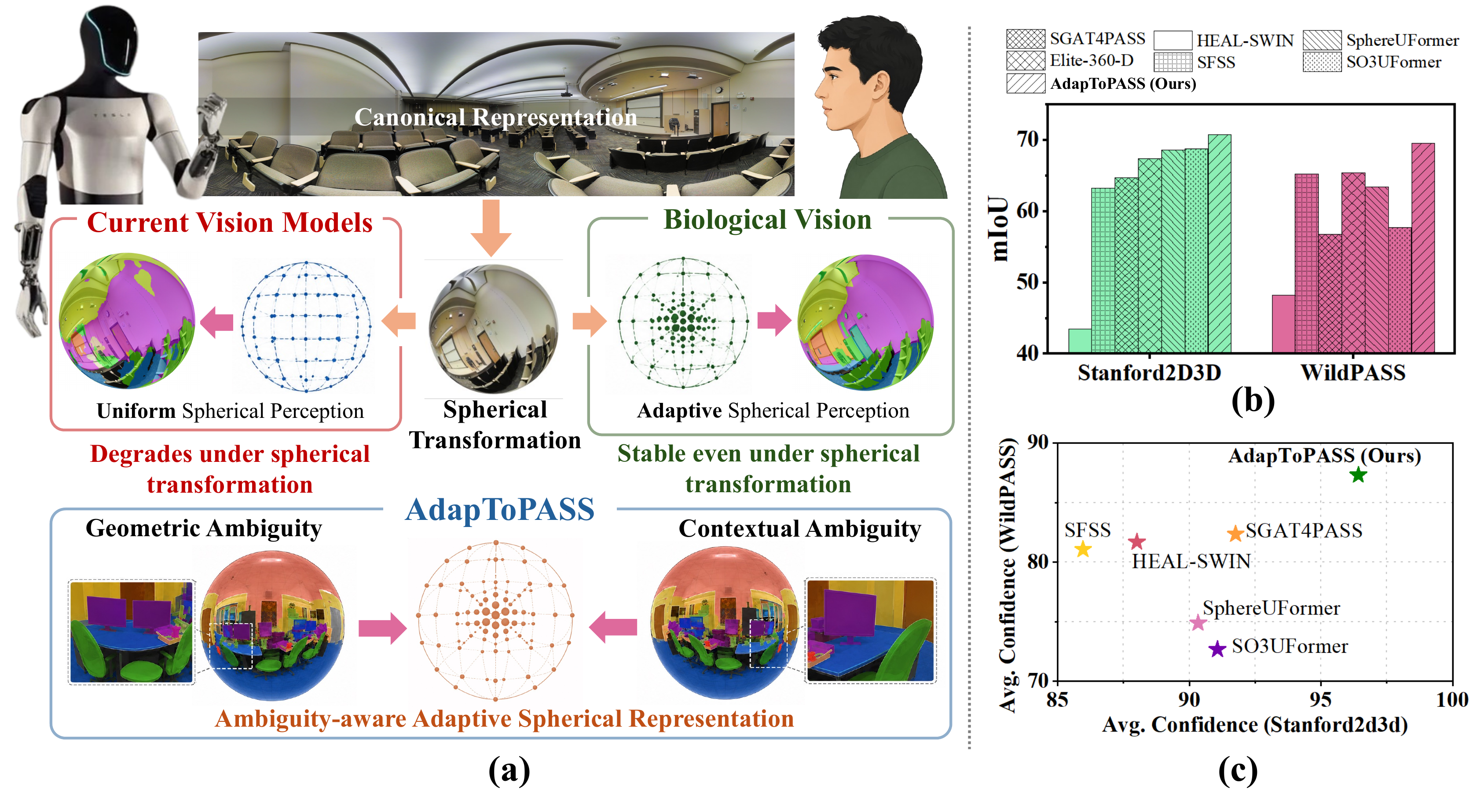}
    \vspace{-10pt}
    \caption{\textbf{(a)} Spherical perception in current methods degrades under spherical transformations due to geometric and contextual ambiguity, while biological vision remains stable via adaptive perception; AdapToPASS models this adaptivity. \textbf{(b)} Improved segmentation performance of AdapToPASS on Stanford2D3D and WildPASS. \textbf{(c)} AdapToPASS exhibits higher and more consistent confidence, indicating stronger geometric and contextual robustness. Click \href{https://empactlab.github.io/AdapToPASS-NeurIPS-2026/}{\textcolor{DeepPink}{\underline{here}}} to visit the project page.}
    \label{fig:teaser}
\end{figure}

\begin{abstract}

Spherical Transformers have emerged as a promising framework for panoramic semantic segmentation (PASS) by operating directly on spherical geometry and alleviating projection-induced distortions.
However, existing architectures rely on assumptions of canonical spherical structure and stable viewpoints, which are frequently violated in real-world $360^\circ$ imagery due to unconstrained camera motion, introducing significant contextual and geometric ambiguity.
Consequently, they lack adaptive mechanisms to model such ambiguity, limiting robustness to unseen spherical transformations.
In contrast, biological perception is inherently ambiguity-aware: rather than estimating uncertainty probabilistically, it adapts to fluctuations in cue reliability caused by geometric and contextual variations, enabling stable interpretation under complex visual transformations.
Motivated by these observations, we first present a \textbf{systematic analysis} of existing PASS architectures under various unseen spherical transformations.
Following this, we introduce \textbf{AdapToPASS}, a \textbf{novel}, \textbf{bio-inspired} Spherical Transformer that adaptively models contextual and geometric ambiguities for robust panoramic semantic segmentation.
At the core of AdapToPASS are the \textbf{Adaptive Spherical Attention (AdaSpA)} blocks that dynamically modulate attention according to local contextual ambiguity, mimicking the adaptive, context-driven perception of biological vision.
To address geometric ambiguities, AdapToPASS employs \textbf{Bifocal Spherical Representation} that reconciles the trade-off between field of view and spatial resolution; along with \textit{boundary supervision} to emulate the boundary-sensitive nature of biological vision.
We evaluate AdapToPASS in both indoor and outdoor semantic segmentation, where it consistently outperforms prior state-of-the-art methods. 
We further validate AdapToPASS under unseen spherical transformations, where it demonstrates strong robustness and surpasses the next-best method by \textbf{+13.38\%} relative improvement in mIoU on \textbf{Stanford2D3D} and \textbf{+18.77\%} on \textbf{WildPASS}.
Additionally, we introduce a lightweight variant, \textbf{AdapToPASS-Swift}, with fewer than \textbf{2M parameters}, which surpasses compact baselines while retaining robustness to spherical transformations.

\end{abstract}

\vspace{-20pt}
\section{Introduction}
\vspace{-6pt}
Panoramic semantic segmentation plays a crucial role in perception for applications such as robotic navigation and virtual reality, where understanding the full $360^\circ$ surrounding environment is essential for decision-making~\cite{ai2022deep, da2023omnidirectional, zhang2023survey, manh2025mind, damodaran2026eagor}. 
Although semantic segmentation has achieved remarkable success in perspective images with Convolutional Neural Networks (CNNs), encoder-decoder architectures, and Transformer-based models~\cite{long2015fully, ronneberger2015u, badrinarayanan2017segnet, zheng2021rethinking, xie2021segformer}, these approaches are inherently designed for perspective 2D images and fail to capture the unique geometric properties of spherical data~\cite{yang2021context}. 
A common workaround is to project spherical images into the equirectangular projection (ERP) for compatibility with standard 2D architectures. However, ERP introduces severe distortions, breaks the left-right continuity at image boundaries, and spans an extremely large field-of-view (FoV). This limits the effectiveness of the 2D architectures~\cite{su2017learning, cohen2018spherical, coors2018spherenet, tateno2018distortion, zhang2024behind}.
To overcome this limitation, several spherical Transformer backbones have been introduced to operate directly on the sphere~\cite{carlsson2024heal, benny2025sphereuformer}. 
Despite their strong performance, these models rely on assumptions of canonical spherical structure and stable viewpoints. 
In practice, however, omnidirectional cameras on handheld devices, and mobile robot platforms undergo unconstrained motion, introducing arbitrary rotations, scale changes, and viewpoint shifts that violate these assumptions and substantially degrade segmentation accuracy~\cite{li2023sgat4pass, zhu2026so3uformer}.

\begin{figure*}[t!]
    \centering
    \includegraphics[width=1.0\textwidth]{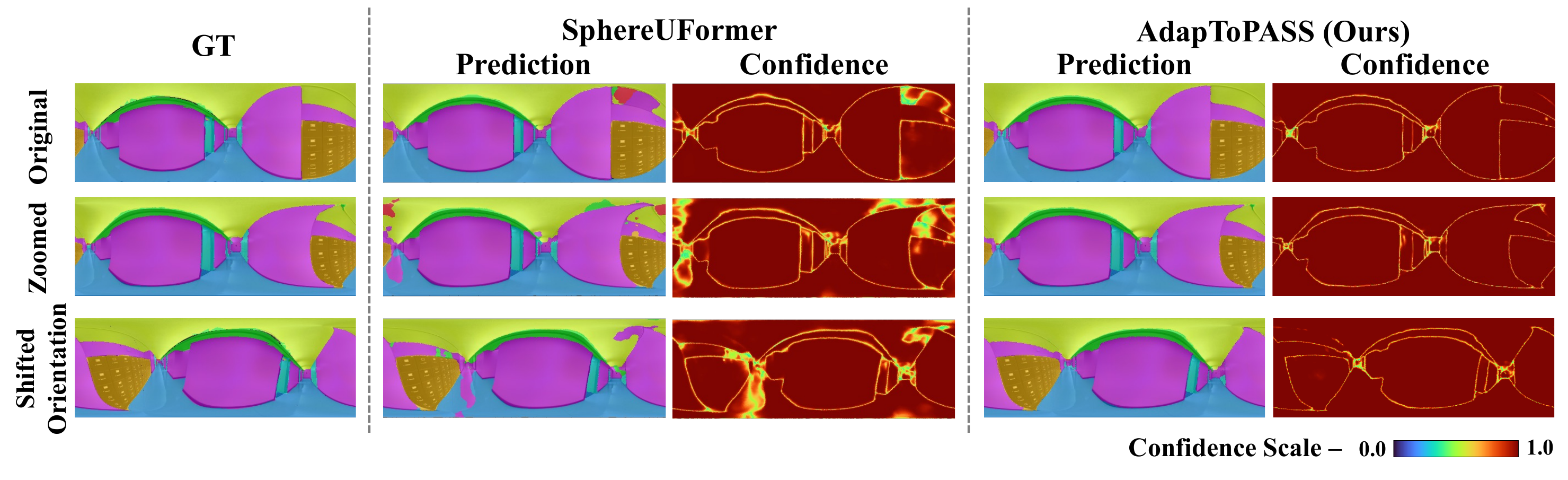}
    \vspace{-20pt}
    \caption{\textbf{Qualitative comparison of AdapToPASS and SphereUFormer on Stanford2D3D under unseen spherical transformations.}
    We show segmentation predictions and confidence maps, where \textbf{confidence is defined as the inverse of ambiguity} \colorbox{orange!10}{$\mathrm{Conf}=1-\mathrm{Amb}$} and visualized using the maximum softmax probability. SphereUFormer shows a clear confidence drop under spherical transformations, whereas AdapToPASS maintains higher confidence and sharper semantic boundaries.}
    \label{fig:confidence_exp}
    \vspace{-15pt}
\end{figure*}

In this work, we first conduct a systematic analysis of commonly used spherical Transformer backbones under a diverse set of spherical transformations, which revealed clear performance degradation patterns (see Fig.~\ref{fig:confidence_exp}). 
More critically, our analysis reveals \textbf{two} fundamental limitations in these architectures: 
\textbf{(1) Contextual ambiguity.} Existing methods rely on uniform feature aggregation regardless of scene complexity, failing to adapt when local regions exhibit ambiguous or overlapping class boundaries (Fig~\ref{fig:teaser}\textcolor{DeepPink}{a}). 
\textbf{(2) Geometric ambiguity.} They treat all spatial locations equally, without the varying degrees of distortion and structural irregularity introduced by spherical transformations.

From a neuroscience perspective, biological vision is inherently adaptive, multi-scale, and ambiguity-aware. 
Rather than processing the entire visual field uniformly, the human visual system dynamically adjusts contextual integration according to local ambiguity, and expands receptive context in uncertain regions while preserving fine detail in reliable ones~\cite{yang2025increasing, vacher2023measuring, debnath2025rasp, debnath2026llmind, liu2026vl2spike}. 
It further balances a trade-off between FoV and spatial resolution to resolve ambiguity across different spatial scales, which enables both global scene understanding and high-acuity local perception~\cite{land2012animal, schwartz1980computational, warrant2004vision}. 
Moreover, biological perception is boundary-aware, as it helps disambiguate object contours and region transitions even under appearance and viewpoint changes~\cite{parrey2024cats}.
All these characteristics of biological vision contribute to their remarkable robustness to unseen visual transformations, including changes in orientation, scale, and viewpoint~\cite{nandakumar2011invariance, han2020scale, rolls2021learning, okamura2025view} (Fig~\ref{fig:teaser}\textcolor{DeepPink}{a}).

These observations raises an interesting question: \textit{\textbf{How can ambiguity-aware perception in human vision be efficiently and adaptively integrated into Spherical Transformers to achieve robust panoramic semantic segmentation?
}} 
To this end, we propose \textbf{AdapToPASS}, a bio-inspired panoramic semantic segmentation framework that models: 
(\textbf{I}) \underline{contextual ambiguity} via \textbf{Adaptive Spherical Attention (AdaSpA)} blocks, which adaptively modulate attention according to local contextual ambiguity; and (\textbf{II})~\underline{geometric ambiguity} via \textit{Bifocal Spherical Representation} with dual pathways, an \textbf{Acuity Stream} focusing on high-resolution detail and a \textbf{Lateral Stream} focusing on wide FoV context, effectively balancing the FoV vs. resolution trade-off.
Additionally, we incorporate \textit{SDF-based boundary supervision} to explicitly capture boundary ambiguity and improve localization.

We evaluate AdapToPASS across both indoor and outdoor panoramic semantic segmentation benchmarks.
We further assess generalization under \textbf{unseen spherical transformations} across varying difficulty levels. AdapToPASS it surpasses the previous best-performing method by \textbf{+13.38\%} \textit{relative improvement in mIoU} on \textbf{Stanford2D3D}~\cite{armeni2017joint} and \textbf{+18.77\%} on \textbf{WildPASS}~\cite{yang2021capturing, yang2021context}.

To summarize, the major contributions of this paper are four-fold:
\vspace{-5pt}
\begin{itemize}[leftmargin=10pt, itemsep=2pt, topsep=2pt, parsep=0pt, partopsep=0pt]
    \item Inspired by neuroscience-grounded principles, we identify a fundamental limitation of existing panoramic segmentation models and conduct a systematic analysis of their generalization under unseen spherical transformations.
    
    \item We propose \textit{AdaSpA} blocks that adaptively modulate attention according to local contextual ambiguity, serving as a computational analogue of uncertainty-driven semantic processing in biological vision (Sec~\ref{sec:AdaSpA}).
    
    \item We introduce the \textit{Bifocal Spherical Representation} framework, inspired by the biological trade-off between spatial acuity and field of view (Sec.~\ref{sec:parallel}), which serves as a computational analogue for addressing geometric ambiguities.
    
    \item We validate our method on both indoor and outdoor panoramic segmentation benchmarks, and evaluate its generalization under unseen spherical transformations (Sec.~\ref{sec:experiment}). We further introduce a compact 2M-parameter variant, showing that ambiguity-aware spherical perception can be preserved under strict efficiency constraints (Sec.~\ref{sec:results}).
\end{itemize}
\vspace{-5pt}
\section{Related Works}
\vspace{-5pt}

Recently, bio-inspired mechanisms have gained widespread adoption across diverse domains including art, robotics and material design~\cite{debnath2024scribgen, sakshi2026robotic, debnath2025computational, debnath2023modified}.

\noindent\textbf{Panoramic Semantic Segmentation.}
For $360^\circ$ imagery, equirectangular projection (ERP) remains the most widely used representation due to its simplicity and compatibility with standard vision architectures. 
However, ERP introduces severe latitude-dependent distortions, especially near the poles, which compromise geometric consistency and semantic reasoning~\cite{su2017learning, Zhang2024GoodSAMBD, cohen2018spherical, Zhang2025Leader360VTL, coors2018spherenet, ai2025survey}. 
Distortion-aware Transformers such as Trans4PASS~\cite{zhang2022bending} and PanoFormer~\cite{shen2022panoformer} mitigate ERP distortions through deformable or sphere-aware tokenization, but remain tied to planar projections.
A more natural alternative is to operate directly on the sphere, to avoid artifacts caused by planar parameterization. 
Foundational efforts such as spherical harmonic networks~\cite{cohen2018spherical, esteves2018learning} and gauge-equivariant CNNs~\cite{cohen2019gauge} established principled frameworks for spherical representation learning.
Building on this direction, recent panoramic segmentation methods increasingly incorporate geometry-aware or native spherical representations. 
HEAL-SWIN~\cite{carlsson2024heal} combines the equal-area HEALPix grid with hierarchical attention;
SFSS~\cite{guttikonda2024single} explores multimodal spherical segmentation using RGB, depth, and normal cues. Similarly, Elite360D~\cite{ai2024elite360d} introduces icosahedron-based bi-projection fusion for efficient panoramic dense perception, and later extended to multi-task panoramic perception and segmentation. 
Most recently, SphereUFormer~\cite{benny2025sphereuformer} proposed a U-shaped Transformer on the icosphere with localized spherical self-attention and geometry-consistent multi-scale processing, achieving state-of-the-art performance in panoramic semantic segmentation.
\textit{Despite these advances, current panoramic segmentation methods largely assume a stable canonical spherical layout during training and evaluation. Their robustness to unseen spherical transformations, including rotations, scale changes, orientation shifts, and viewpoint shifts, remains insufficiently studied}.

\noindent\textbf{Transformation Robustness in Panoramic Segmentation.}
Generalization beyond canonical spherical layouts is critical for real-world panoramic perception.
Most panoramic segmentation methods implicitly assume a canonical camera orientation~\cite{zhu2026so3uformer}. 
This assumption is rarely satisfied in real-world $360^\circ$ imagery, where camera motion and viewpoint changes can induce substantial spherical distortions. 
Recent efforts have begun to address this fragility via transformation-aware modeling. 
For example, prior works have explored M\"obius transformations for data augmentation~\cite{zhou2021data, chhipa2024mobius}, while PanDA~\cite{cao2025panda} uses them to enforce consistency under geometric deformations in panoramic depth estimation. 
M\"obius-equivariant CNNs~\cite{mitchel2022mobius} further provide a theoretical foundation for conformal-equivariant spherical CNNs. 
However, these works do not explicitly study robustness to diverse spherical transformations in panoramic segmentation. 
Within panoramic segmentation, existing efforts primarily focus only on SO(3) rotations or rely on transformation-specific augmentation. SGAT4PASS~\cite{li2023sgat4pass} improves robustness to 3D disturbances through spherical geometry-aware modeling, while the concurrent SO3UFormer~\cite{zhu2026so3uformer} enhances rotation robustness using gauge-aware relative positional mechanisms and data augmentation. 
More recently, Unified Spherical Frontend~\cite{yu2025unified} achieves cross-camera rotation equivariance by construction through distance-only convolution kernels on icosahedral grids.
\textit{However, these methods mainly improve geometric consistency under specific perturbations, rather than providing an adaptive mechanism for broader transformation-induced ambiguity.}
\vspace{-5pt}

\paragraph{Geometric and Context Modeling for Segmentation.}
Panoramic semantic segmentation is inherently ambiguous: wide-field imagery introduces unreliable long-range cues, spherical distortions, viewpoint changes, and unstable semantic transitions, leading to contextual, geometric, and boundary ambiguity.
In particular, Dinh et al.~\cite{cao2024geometric} show that contextual ambiguity in monocular $360^\circ$ imagery limits uniform feature aggregation, highlighting the need for adaptive context modeling.
Existing panoramic methods incorporate structured contextual aggregation~\cite{yang2021context,hu2021new} or identify semantic/geometric ambiguity in distortion-heavy regions~\cite{duan2025panoramicoutofdistributionsegmentation}, while geometric-aware models mitigate projection-induced distortions through sphere-aware tokenization or spherical representations~\cite{shen2022panoformer, carlsson2024heal}. 
Boundary-aware segmentation losses further show that explicitly modeling semantic transitions can reduce diffuse predictions near object boundaries~\cite{wang2022active}.
Beyond segmentation, adaptive models such as HAFA~\cite{chen2023building} and AMContrast3D~\cite{chen2024adaptive} demonstrate the value of uncertainty-guided feature modulation.
\textit{However, these efforts do not provide a unified mechanism for jointly resolving contextual, geometric, and boundary ambiguity within a native segmentation architecture}.
\begin{center}
\vspace{-6pt}
\colorbox{orange!10}{
\begin{minipage}{0.96\linewidth}
    \textbf{Takeaway.} 
    To our knowledge, this is \textit{the first} unified architectures to jointly model contextual, geometric, and boundary ambiguity for panoramic semantic segmentation. 
    AdapToPASS introduces an ambiguity-aware Spherical Transformer that adaptively aggregates context, handles spherical geometric distortion, and predicts sharper semantic boundaries directly on the sphere. 
\end{minipage}
}
\end{center}
\begin{figure*}[t!]
    \centering
    \includegraphics[width=0.94\textwidth]{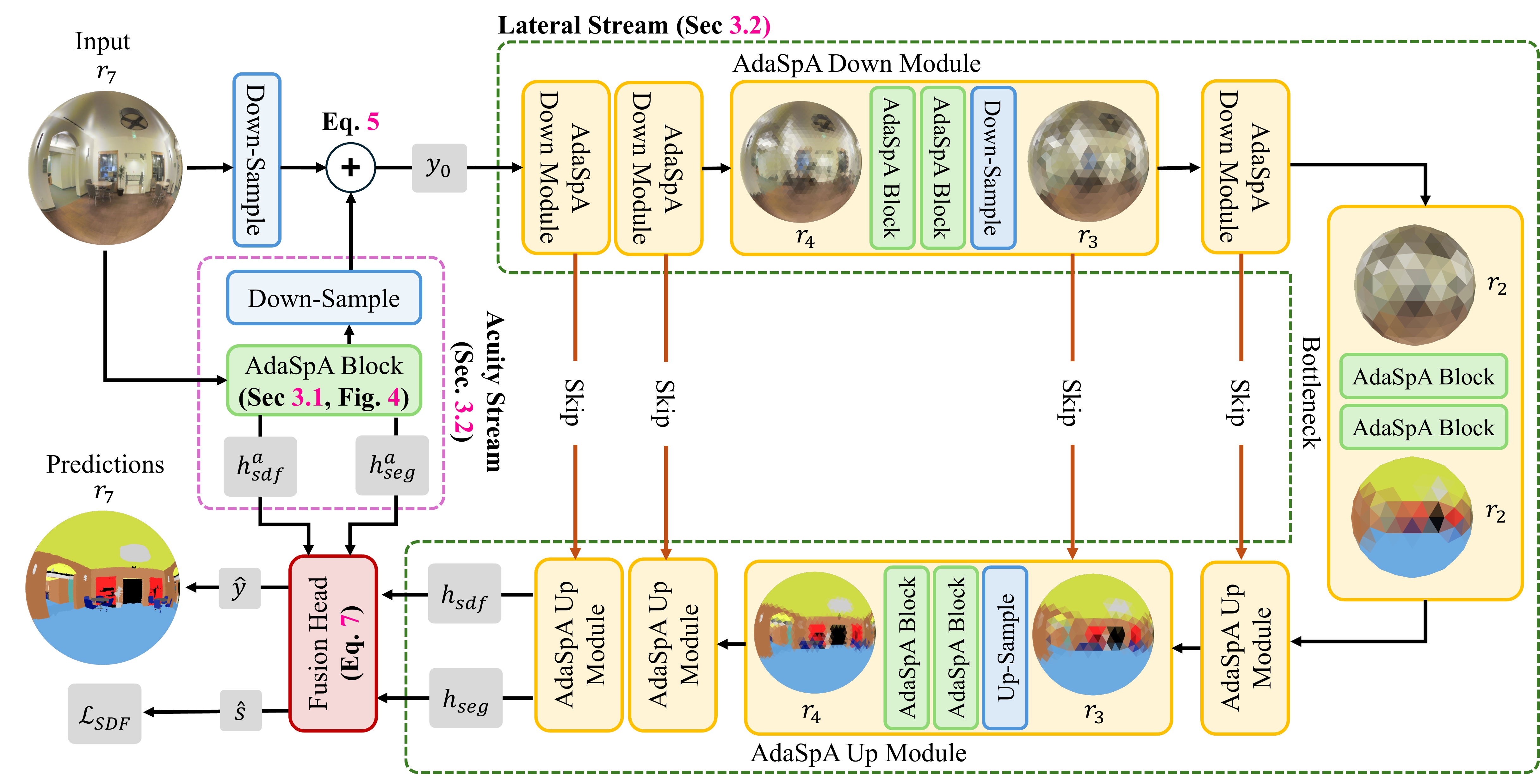}
    \vspace{-2pt}
    \caption{Overview of the AdapToPASS framework. \textit{Zoom in for a better view}.}
    \label{fig:methodology}
    \vspace{-15pt}
\end{figure*}

\vspace{-8pt}
\section{Methodology}
\label{sec:full_methods}
\vspace{-7pt}
\paragraph{Overview.}
An overview of AdapToPASS is illustrated in Fig.~\ref{fig:methodology}. 
Given an omnidirectional RGB image, we first projects it onto an icosphere, obtaining a spherical signal $\mathbf{x} \in \mathbb{R}^{N \times 3}$ over $N$ icosphere vertices.
The model produces per-node semantic logits $\hat{\mathbf{y}} \in \mathbb{R}^{N \times K}$ and per-class signed distance field (SDF) predictions $\hat{\mathbf{s}} \in \mathbb{R}^{N \times K}$, where $K$ is the number of semantic classes.
Additional implementation details are provided in Appendix~\ref{apps:icosphere_repp}.

Conceptually, we formulate AdapToPASS from an energy-minimization perspective:
{\setlength\abovedisplayskip{3pt}
\setlength\belowdisplayskip{3pt}
\begin{equation}
(\hat{\mathbf{Y}},\hat{\mathbf{S}})
=
\arg\min_{\mathbf{Y},\mathbf{S}}
\mathcal{E}_{\mathrm{sem}}(\mathbf{Y};\mathbf{X})
+
\lambda_{c}\,\mathcal{E}_{\mathrm{ctx}}(\mathbf{Y};\mathbf{c},\mathcal{G})
+
\lambda_{g}\,\mathcal{E}_{\mathrm{geo}}(\mathbf{Y},\mathbf{S};\mathbf{D}^{\partial},\mathcal{G})
\end{equation}}
\noindent This formulation estimates semantic predictions $\hat{\mathbf{Y}}$ and SDF predictions $\hat{\mathbf{S}}$ by minimizing three complementary energies: semantic prediction $\mathcal{E}_{\mathrm{sem}}$; \underline{contextual ambiguity} $\mathcal{E}_{\mathrm{ctx}}$ modeled by AdaSpA through the learned ambiguity signal $c_i$ and the induced contextual-ambiguity-aware geodesic bias $b_{ij}^h(c_i)$ (Sec.~\ref{sec:AdaSpA}); and \underline{geometric ambiguity} $\mathcal{E}_{\mathrm{geo}}$ modeled by the Bifocal Spherical Representation Framework (Sec.~\ref{sec:parallel}). Here, $\mathbf{c}$ controls adaptive context aggregation, while $\mathbf{D}^{\partial}$ provides geometric boundary guidance on the icosphere graph $\mathcal{G}$.

\subsection{Adaptive Spherical Attention (AdaSpA) Blocks}
\label{sec:AdaSpA}

\begin{wrapfigure}{l}{0.7\columnwidth}
\vspace{-15pt}
\centering
\includegraphics[width=\linewidth]{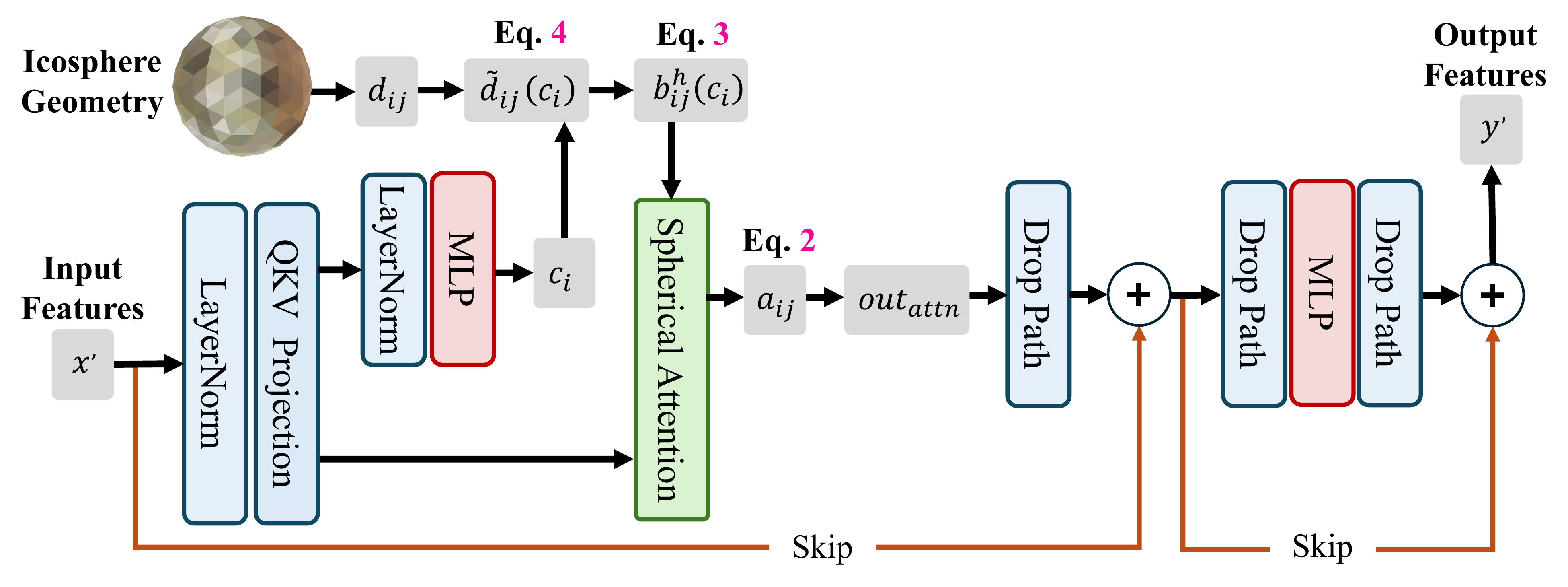}
\vspace{-20pt}
\caption{Overview of an AdaSpA Block.}
\label{fig:adaspa}
\vspace{-10pt}
\end{wrapfigure}

\textit{\textbf{Insight.}}
Standard spherical attention mechanisms aggregate context uniformly across the sphere, assigning equal weight to all neighbors regardless of their local contextual ambiguity~\cite{benny2025sphereuformer, carlsson2024heal}.
When sensory evidence is ambiguous, human perception relies more on context and priors; when reliable, it follows the local signal.
We propose Adaptive Spherical Attention (AdaSpA), which dynamically modulates contextual integration by adapting each token's attention to a learned local contextual ambiguity signal. The overview of the block is illustrated in Fig.~\ref{fig:adaspa}.
\vspace{-5pt}
\paragraph{Spherical Attention.}
At each icosphere rank $r$, an AdaSpA block performs local self-attention over the $w$-hop neighborhood $\mathcal{N}_{w}(i)$ of each node $v_i$.
Given token features $\mathbf{x}\in\mathbb{R}^{N\times d}$, queries, keys, and values are computed from linear projections of neighboring tokens.
For head $h$, the attention logit between node $v_i$ and its neighbor $v_j$ is defined as
{\setlength\abovedisplayskip{3pt}
\setlength\belowdisplayskip{3pt}
\begin{equation}
    a_{ij}^{h}
    =
    \tau_h
    \hat{\mathbf{Q}}_{i}^{h}
    \hat{\mathbf{K}}_{j}^{h\top}
    +
    b_{ij}^{h}(c_i),
    \label{eq:attn}
\end{equation}}
where $\hat{\mathbf{Q}}$ and $\hat{\mathbf{K}}$ are $\ell_2$-normalized queries and keys, $\tau_h$ is a learnable head-specific scale, and $b_{ij}^{h}(c_i)$ is an ambiguity-aware geodesic bias.
This formulation retains standard local self-attention while injecting spherical geometry directly into the attention logits.
\vspace{-5pt}
\paragraph{Geodesic Bias.}
To encode the intrinsic geometry of the sphere, we compute the geodesic distance between neighboring icosphere nodes:
$
    d_{ij}
    =
    \arccos\!\left(\mathbf{n}_{i}^{\top}\mathbf{n}_{j}\right),
    \label{eq:geo_dist}
$
where $\mathbf{n}_i$ and $\mathbf{n}_j$ are unit sphere normals.
Rather than relying only on learned relative position embeddings, we map this distance to a head-specific attention bias:
{\setlength\abovedisplayskip{3pt}
\setlength\belowdisplayskip{3pt}
\begin{equation}
    b_{ij}^{h}
    =
    \lambda_{\mathrm{geo}}\,
    \phi_h(d_{ij}),
    \label{eq:geo_bias}
\end{equation}}
where $\phi_h(\cdot)$ is a lightweight MLP producing the bias for head $h$, and $\lambda_{\mathrm{geo}}$ is a learnable scale.
This encourages attention to respect spherical proximity while remaining learnable.
\vspace{-5pt}
\paragraph{Contextual-Ambiguity-Aware Geodesic Bias.}
A fixed geodesic bias applies the same spatial preference to all tokens, regardless of local semantic difficulty.
However, ambiguous regions often require broader contextual evidence, whereas confident regions can rely more strongly on local structure.
AdaSpA therefore predicts a contextual ambiguity score $c_i\in[0,1]$ for each query token and uses it to rescale the geodesic distance:
\begin{equation}
    \tilde{d}_{ij}(c_i)
    =
    \operatorname{clamp}
    \left(
    \frac{d_{ij}}
    {s_{\min} + (s_{\max}-s_{\min})c_i + \varepsilon},
    0,\pi
    \right).
    \label{eq:uagb_scale}
\end{equation}
The modulated distance $\tilde{d}_{ij}(c_i)$ replaces $d_{ij}$ in Eq.~\eqref{eq:geo_bias}, yielding the ambiguity-aware bias $b_{ij}^{h}(c_i)$ used in Eq.~\eqref{eq:attn}.
When $c_i$ is high, the effective distance profile is compressed, encouraging broader contextual aggregation; when $c_i$ is low, the bias remains sharper and more local.
Each AdaSpA block follows a standard pre-normalization attention-FFN structure.
\vspace{-5pt}


\subsection{Bifocal Spherical Representation Framework}
\label{sec:parallel}
\vspace{-5pt}
\textit{\textbf{Insight.}}
Conventional spherical segmentation models generally follow a single hierarchical pathway, where expanding the receptive field progressively reduces spatial resolution~\cite{benny2025sphereuformer, zhu2026so3uformer}.
This creates a fundamental tension between acuity and field of view.
Inspired by biological vision, where foveated eyes prioritize high spatial acuity while laterally placed eyes favor wide-field contextual awareness, AdapToPASS introduces a \textit{Bifocal Spherical Representation Framework} with two specialized streams.
The \textbf{Acuity Stream} focus on the full-resolution icosphere ($r{=}7$) to preserve local geometry and boundary detail, and the \textbf{Lateral Stream} focus on a downsampled icosphere ($r'{=}r{-}1{=}6$) to aggregate broader semantic context.

\noindent\textbf{Acuity Stream.}
The Acuity Stream preserves fine-grained spherical structure that may be weakened by hierarchical downsampling in the Lateral Stream.
Given the full-resolution icosphere input $\mathbf{x}\in\mathbb{R}^{N\times3}$, we first project it into a lightweight feature space and process it with an AdaSpA block, which produce full-resolution acuity features $\mathbf{F}_{\mathrm{acuity}}\in\mathbb{R}^{N\times d_a}$.
These features are injected into the Lateral Stream through a gated projection:
{\setlength\abovedisplayskip{3pt}
\setlength\belowdisplayskip{3pt}
\begin{equation}
    \mathbf{y}_{0}
    =
    \mathbf{E}(\mathbf{x})
    +
    \sigma(g_{\mathrm{gate}})\,
    \Pi(\mathbf{F}_{\mathrm{acuity}}),
    \label{eq:acuity_widefov_fusion}
\end{equation}}
where $\mathbf{E}(\cdot)$ denotes the Lateral input projection and $\Pi(\cdot)$ downsamples and projects the acuity features to the Lateral resolution.
To retain full-resolution supervision, the Acuity Stream also produces auxiliary segmentation and SDF predictions. 

\noindent \textbf{Lateral Stream.}
The Lateral Stream is designed to capture broad semantic and geometric context under a larger effective field of view.
Following the U-shaped spherical hierarchy of SphereUFormer~\cite{benny2025sphereuformer}, it operates on the downsampled icosphere at rank $r'{=}r{-}1$.
Given the fused input feature $\mathbf{y}_0$ from Eq.~\ref{eq:acuity_widefov_fusion}, the stream first encodes features through a sequence of Adaptive Spherical Attention Down Modules (AdaSpA Down Modules) \textit{encoders} with progressive graph \textit{downsampling}, thereby increasing the receptive field while reducing spatial resolution.
A \textit{bottleneck} stage at the coarsest icosphere resolution aggregates global scene-level context.
A series of AdaSpA Up Modules \textit{decoders} then progressively upsamples the features and combines them with encoder features through skip connections:
{\setlength\abovedisplayskip{3pt}
\setlength\belowdisplayskip{3pt}
\begin{equation}
    \mathbf{y}_{i}^{\mathrm{dec}}
    =
    f_{i}^{\mathrm{dec}}
    \!\left(
    \left[
    \mathrm{LN}(\mathbf{y}_{i}^{\mathrm{up}}),
    \mathrm{LN}(\mathbf{e}_{S-1-i})
    \right]
    \right),
    \label{eq:skip}
\end{equation}}
where $\mathbf{y}_{i}^{\mathrm{dec}}$ is the decoder feature at stage $i$, $\mathbf{y}_{i}^{\mathrm{up}}$ is the upsampled feature, and $\mathbf{e}_{S-1-i}$ is the corresponding encoder skip feature. 
Here, $[\cdot,\cdot]$ denotes channel-wise concatenation, $\mathrm{LN}$ denotes layer normalization, and $f_i^{\mathrm{dec}}$ is the decoder block.
The resulting feature $\mathbf{y}_{L}\in\mathbb{R}^{N'\times d}$ is finally lifted back to the full icosphere rank $r$ and fused with the Acuity Stream for dense prediction.

\textbf{Acuity-Latent Fusion Head.} The final outputs combine Lateral and Acuity predictions through learned gates:
{\setlength\abovedisplayskip{3pt}
\setlength\belowdisplayskip{3pt}
\begin{equation}
\begin{aligned}
    \hat{\mathbf{y}} &= \mathbf{h}_{\text{seg}}(\mathbf{y}_{L}) + \sigma(g_{\text{seg}})\,\mathbf{h}_{\text{seg}}^{a}(\mathbf{F}_{\text{acuity}}), \qquad
    \hat{\mathbf{s}} = \mathbf{h}_{\text{sdf}}(\mathbf{y}_{L}) + \sigma(g_{\text{sdf}})\,\mathbf{h}_{\text{sdf}}^{a}(\mathbf{F}_{\text{acuity}})
\end{aligned}
\label{eq:out_heads}
\end{equation}}
where $\hat{\mathbf{y}}\in\mathbb{R}^{N\times K}$ denotes the final semantic logits used for segmentation prediction, $\hat{\mathbf{s}}\in\mathbb{R}^{N\times K}$ denotes the predicted per-class SDF used to compute $\mathcal{L}_{\mathrm{SDF}}$, $\mathbf{y}_{L}$ is the final Lateral decoder feature, and $\mathbf{h}^{a}_{\mathrm{seg}}$ and $\mathbf{h}^{a}_{\mathrm{sdf}}$ denote the Acuity Stream prediction heads.

\vspace{-5pt}
\paragraph{Signed Distance Field (SDF) based Boundary Supervision.}
\label{sec:sdf}
Region-level cross-entropy supervision does not penalize boundary ambiguities, leading the model free to produce diffuse predictions at semantic transitions~\cite{wang2022active}.
Biological perception is strongly boundary-aware, maintaining sharp delineations between regions even under viewpoint changes.
We thus introduce SDF-based boundary supervision, which encodes semantic boundary on the icosphere graph through per-class signed distance targets used as auxiliary supervision during training.

Given semantic labels $\mathbf{y}\in\{0,\ldots,K\}^{N}$, we identify boundary nodes and compute the graph distance $d_i^{\partial}$ from each node to its nearest boundary,
where negative values denote nodes inside the class region and positive values denote outside.
We attach lightweight SDF heads to the Lateral and Acuity features, and supervise them with a truncated regression loss:
{\setlength\abovedisplayskip{3pt}
\setlength\belowdisplayskip{3pt}
\begin{equation}
    \mathcal{L}_{\mathrm{SDF}}
    =
    \frac{1}{|\mathcal{M}|}
    \sum_{(i,k)\in\mathcal{M}}
    \min\!\left(
    |\hat{s}_{i}^{k}-s_{i}^{k,\mathrm{gt}}|,
    \delta
    \right),
    \label{eq:sdf_loss}
\end{equation}}
where $\mathcal{L}_{\mathrm{SDF}}$ is the truncated SDF regression loss, $\mathcal{M}$ is the set of valid non-background node-class pairs, $\hat{s}_{i}^{k}$ and $s_{i}^{k,\mathrm{gt}}$ are the predicted and target signed distances, and $\delta$ is the truncation threshold.

We further use a sign-alignment loss to encourage consistency between semantic predictions and the SDF-induced inside-outside structure:
{\setlength\abovedisplayskip{3pt}
\setlength\belowdisplayskip{3pt}
\begin{equation}
    \mathcal{L}_{\mathrm{align}}
    =
    \frac{1}{|\mathcal{M}|}
    \sum_{(i,k)\in\mathcal{M}}
    \left|
    \operatorname{softmax}(\hat{\mathbf{y}}_i)_k
    -
    \mathbb{1}[\hat{s}_{i}^{k}<0]
    \right|,
    \label{eq:sdf_align}
\end{equation}}
where $\mathcal{L}_{\mathrm{align}}$ is the alignment loss, $\hat{\mathbf{y}}_i\in\mathbb{R}^{K}$ is the semantic logit vector at node $i$, 
and $\mathbb{1}[\hat{s}_{i}^{k}<0]$ is the indicator that the predicted SDF places node $i$ inside class $k$. 
\subsection{Training Objective and Adaptive Inference}
\vspace{-5pt}

AdapToPASS is trained with a joint objective that combines semantic segmentation, boundary-aware SDF supervision, semantic-SDF consistency, and ambiguity regularization:
{\setlength\abovedisplayskip{3pt}
\setlength\belowdisplayskip{3pt}
\begin{equation}
\label{eq:obj}
\mathcal{L}_{\mathrm{overall}}
=
\mathcal{L}_{\mathrm{CE}}
+
\lambda_{\mathrm{Lov}}\,\mathcal{L}_{\mathrm{Lov}}
+
\lambda_{\mathrm{SDF}}\,\mathcal{L}_{\mathrm{SDF}}(\hat{\mathbf{S}},\mathbf{S}^{\mathrm{gt}}(\mathbf{D}^{\partial},\mathbf{Y}))
+
\lambda_{\mathrm{align}}\,\mathcal{L}_{\mathrm{align}}(\hat{\mathbf{Y}},\hat{\mathbf{S}})
+
\lambda_{c}\,\bar{c}.
\end{equation}}
Here, $\mathcal{L}_{\mathrm{CE}}$ and $\mathcal{L}_{\mathrm{Lov}}$ supervise semantic prediction, $\mathcal{L}_{\mathrm{SDF}}$ provides boundary-aware geometric supervision, $\mathcal{L}_{\mathrm{align}}$ encourages consistency between semantic logits and predicted SDF structure, and $\bar{c}$ regularizes the learned ambiguity signal. Detailed formulations are provided in Appendix~\ref{apps:training_objective}.

\textbf{AdapToPASS remains adaptive at during inference} because the contextual ambiguity signal is computed within the forward pass rather than introduced as a training-only auxiliary variable.
For each input panorama, AdaSpA predicts a token-wise ambiguity score $c_i$ from the current spherical feature $\mathbf{x}_i$ and uses it to modulate the geodesic distance scale in Eq.~\eqref{eq:uagb_scale}.
This yields an input-dependent attention bias $b_{ij}^{h}(c_i)$, allowing each token to adjust its effective aggregation range on the sphere without test-time optimization or gradient updates.
\vspace{-10pt}
\section{Experiments and Evaluations}
\label{sec:experiment}

\begin{table*}[t]
\centering
\small
\resizebox{\textwidth}{!}{%
\begin{tabular}{l 
>{\columncolor{green!6}}c >{\columncolor{green!6}}c >{\columncolor{green!6}}c >{\columncolor{green!6}}c >{\columncolor{green!6}}c
>{\columncolor{orange!8}}c >{\columncolor{orange!8}}c >{\columncolor{orange!8}}c >{\columncolor{orange!8}}c >{\columncolor{orange!8}}c
>{\columncolor{red!6}}c >{\columncolor{red!6}}c >{\columncolor{red!6}}c >{\columncolor{red!6}}c >{\columncolor{red!6}}c}
\toprule
\multirow{2}{*}{\textbf{Method}} 
& \multicolumn{5}{c}{\cellcolor{green!15}\textbf{Mild}} 
& \multicolumn{5}{c}{\cellcolor{orange!18}\textbf{Robust}} 
& \multicolumn{5}{c}{\cellcolor{red!15}\textbf{Stress}} \\
\cmidrule(lr){2-6} \cmidrule(lr){7-11} \cmidrule(lr){12-16}
& \textbf{Rot.} & \textbf{Scale} & \textbf{\shortstack{Ori.}} & \textbf{Trans.} & \textbf{\shortstack{View.}}
& \textbf{Rot.} & \textbf{Scale} & \textbf{\shortstack{Ori.}} & \textbf{Trans.} & \textbf{\shortstack{View.}}
& \textbf{Rot.} & \textbf{Scale} & \textbf{\shortstack{Ori.}} & \textbf{Trans.} & \textbf{\shortstack{View.}} \\
\midrule
SFSS          & 50.40 & 55.47 & 47.02 & 59.45 & 48.51 & 35.65 & 47.84 & 34.14 & 52.84 & 33.00 & 24.66 & 42.39 & 24.20 & 46.11 & 20.26 \\
SGAT4PASS     & 59.92 & 62.62 & 56.27 & 65.05 & 57.19 & 37.59 & 55.83 & 36.11 & 56.64 & 32.61 & 18.56 & 49.22 & 18.74 & 47.68 & 12.50 \\
\midrule
Elite360D     & 54.21 & 58.92 & 48.23 & 64.17 & 51.51 & 31.37 & 47.02 & 28.08 & 56.00 & 29.26 & 17.01 & 38.12 & 15.05 & 47.77 & 14.52 \\
HEAL-SWIN     & 36.94 & 40.71 & 33.13 & 43.61 & 32.18 & 25.42 & 34.08 & 24.23 & 34.60 & 20.54 & 17.01 & 30.39 & 15.61 & 28.65 & 12.40 \\
SphereUFormer & 64.01 & 69.33 & 59.16 & 67.23 & 55.27 & 36.29 & 60.13 & 33.89 & 57.21 & 30.76 & 18.96 & 51.93 & 17.61 & 48.29 & 14.66 \\
SO3UFormer    & 65.85 & 65.30 & 60.01 & 66.70 & 59.81 & 46.88 & 55.11 & 41.95 & 53.94 & 36.64 & 27.07 & 46.53 & 24.01 & 41.70 & 19.19 \\
\midrule
\rowcolor{gray!12}
\textbf{AdapToPASS (Ours)}   
& \textbf{70.93} & \textbf{71.44} & \textbf{66.12} & \textbf{72.43} & \textbf{65.90}
& \textbf{51.52} & \textbf{62.23} & \textbf{46.61} & \textbf{62.90} & \textbf{45.20}
& \textbf{30.43} & \textbf{54.10} & \textbf{27.05} & \textbf{53.17} & \textbf{25.84} \\
\bottomrule
\end{tabular}%
}
\caption{\textbf{mIoU comparison under unseen spherical transformations on the Stanford2D3D benchmark.} Results are reported across mild, robust, and stress settings, including rotation (Rot.), Scale, Orientation Shifts (Ori.), Translation (Trans.), and Viewpoint Shifts (View.). \textit{Higher is better}.}
\label{tab:robustness_results_stanford2d3d}
\vspace{-15pt}
\end{table*}

\vspace{-5pt}
\subsection{Settings and Implementation Details}

\vspace{-5pt}
\noindent\textbf{Evaluation and Experimental Settings.}
We evaluate AdapToPASS on two panoramic semantic segmentation benchmarks: \textit{Stanford2D3D}~\cite{armeni2017joint} and \textit{WildPASS}~\cite{yang2021capturing, yang2021context} (\textit{80:10:10} for \textit{train:val:test}). 
Stanford2D3D is evaluated using all semantic classes, following the full indoor scene parsing setting. 

For WildPASS, we evaluate on the three most frequent foreground classes, which together cover approximately $92\%$ of the annotated labels. 
This reduced WildPASS label space is used because our primary objective is to study robustness and transformation invariance under spherical perturbations, rather than recognition of rarely occurring small objects. 
We provide further discussion of this choice in the Appendix~\ref{app:wildpass_reduced}.
All experiments were conducted on a system equipped with 4$\times$ NVIDIA GeForce RTX 5090 GPUs (32 GB VRAM each). The hyperparameter details are provided in Appendix~\ref{apps:hyperparam_adaptopass}.

\begin{wraptable}{r}{0.65\textwidth}
    \vspace{-8pt}
    \centering
    \resizebox{0.64\textwidth}{!}{%
    \begin{tabular}{l|c|ccc|ccc}
        \toprule
        \multirow{2}{*}{\textbf{Model}} 
        & \textbf{Params} 
        & \multicolumn{3}{c|}{\textbf{Stanford2D3D}} 
        & \multicolumn{3}{c}{\textbf{WildPASS}} \\
        
        & \textbf{(M)} 
        & \textbf{mAcc.\,$\uparrow$} 
        & \textbf{mIoU\,$\uparrow$} 
        & \textbf{Conf.\,$\uparrow$}
        & \textbf{mAcc.\,$\uparrow$} 
        & \textbf{mIoU\,$\uparrow$} 
        & \textbf{Conf.\,$\uparrow$} \\
        
        \midrule
        SFSS            & 15.28 & 74.22 & 63.29 & 85.97 & 77.85 & 65.27 & 81.06 \\
        SGAT4PASS       & 15.22 & 73.13 & 64.75 & 91.76 & 72.15 & 56.77 & 82.35 \\
        \midrule
        Elite360D       & 15.41 & 76.57 & 67.42 & 88.10 & 74.36 & 65.40 & 59.65 \\
        HEAL-SWIN       & 14.06 & 72.60 & 43.46 & 88.01 & 64.64 & 48.20 & 81.70 \\
        SphereUFormer   & 14.77 & 78.13 & 68.60 & 90.34 & 79.15 & 63.38 & 74.89 \\
        SO3UFormer      & 14.92 & 78.68 & 68.79 & 91.07 & 72.11 & 57.74 & 72.68 \\
        \midrule
        \rowcolor{gray!10} 
        \textbf{AdapToPASS (Ours)} 
        & 14.78 
        & \textbf{83.08} 
        & \textbf{70.80} 
        & \textbf{96.42}
        & \textbf{81.44} 
        & \textbf{69.58} 
        & \textbf{87.31} \\
        
        \bottomrule
    \end{tabular}%
    }
    \vspace{-4pt}
    \caption{\textbf{Comparison with state-of-the-art panoramic semantic segmentation methods on Stanford2D3D and WildPASS.} We report model parameters, mean accuracy (mAcc), mean IoU (mIoU), and mean prediction confidence (Conf.).}
    \label{tab:sota_comparison}
    \vspace{-10pt}
\end{wraptable}

\noindent \textbf{Unseen Spherical Transformations.}
To evaluate robustness beyond canonical panoramic views, we construct transformed evaluation variants of Stanford2D3D and WildPASS using diverse spherical transformations.
For a fair evaluation of robustness, \textit{all methods are trained only on the original, untransformed training sets}, without any spherical-transformation-based data augmentation. 
The transformed variants are used exclusively for evaluation.
We consider five transformation families: \textit{Rotation}, \textit{Scale}, \textit{Translation}, \textit{Orientation Shift}, and \textit{Viewpoint Shift}. 
Each family is further divided into three difficulty regimes: \textit{mild}, \textit{robust}, and \textit{stress}, with increasingly larger geometric perturbations. 
Further details are provided in Appendix~\ref{app:unseen_spherical_transformations}.
\vspace{-5pt}

\noindent \textbf{Baseline Methods.}
We compare AdapToPASS with representative ERP-based, spherical, and hybrid panoramic segmentation methods. 
For ERP baselines, we include SFSS as a strong RGB-only method and SGAT4PASS~\cite{li2023sgat4pass} as an ERP-based approach with spherical geometry-aware modeling. 
For spherical baselines, we compare with HEAL-SWIN~\cite{carlsson2024heal}, SphereUFormer~\cite{benny2025sphereuformer}, and SO3UFormer~\cite{zhu2026so3uformer}, which operate directly on spherical discretizations. 
We further include Elite360D~\cite{ai2024elite360d}, a hybrid method combining ERP features with a low-resolution icosahedral projection. 
Additional implementation details are provided in Appendix~\ref{apps:baseline_implementation}.

\vspace{-5pt}
\subsection{Results and Discussion}
\vspace{-5pt}
\label{sec:results}

\begin{figure*}[t!]
    \centering
    \begin{subfigure}[t]{0.66\textwidth}
        \centering
        \includegraphics[width=1.0\textwidth]{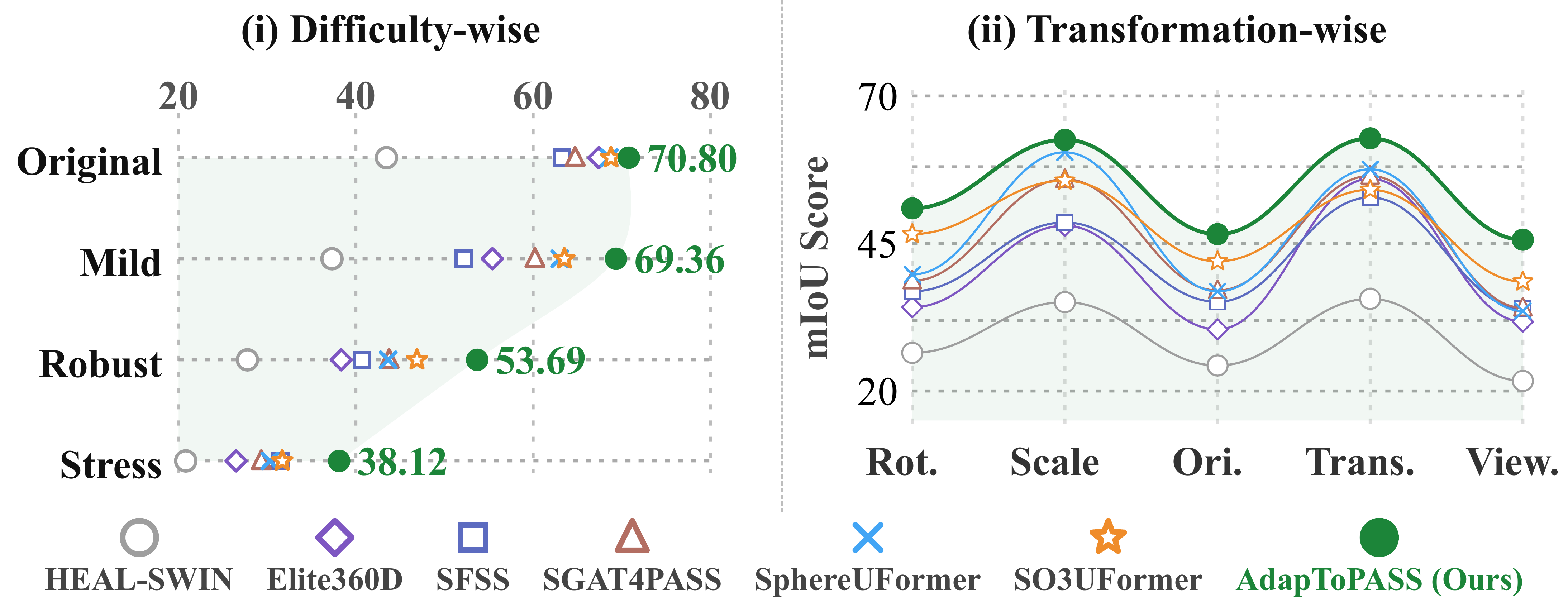}
        \vspace{-15pt}
        \caption{}
        \label{fig:analysis_stanford2d3d}
    \end{subfigure}
    \hfill
    \begin{subfigure}[t]{0.32\textwidth}
        \centering
        \includegraphics[width=1.0\textwidth]{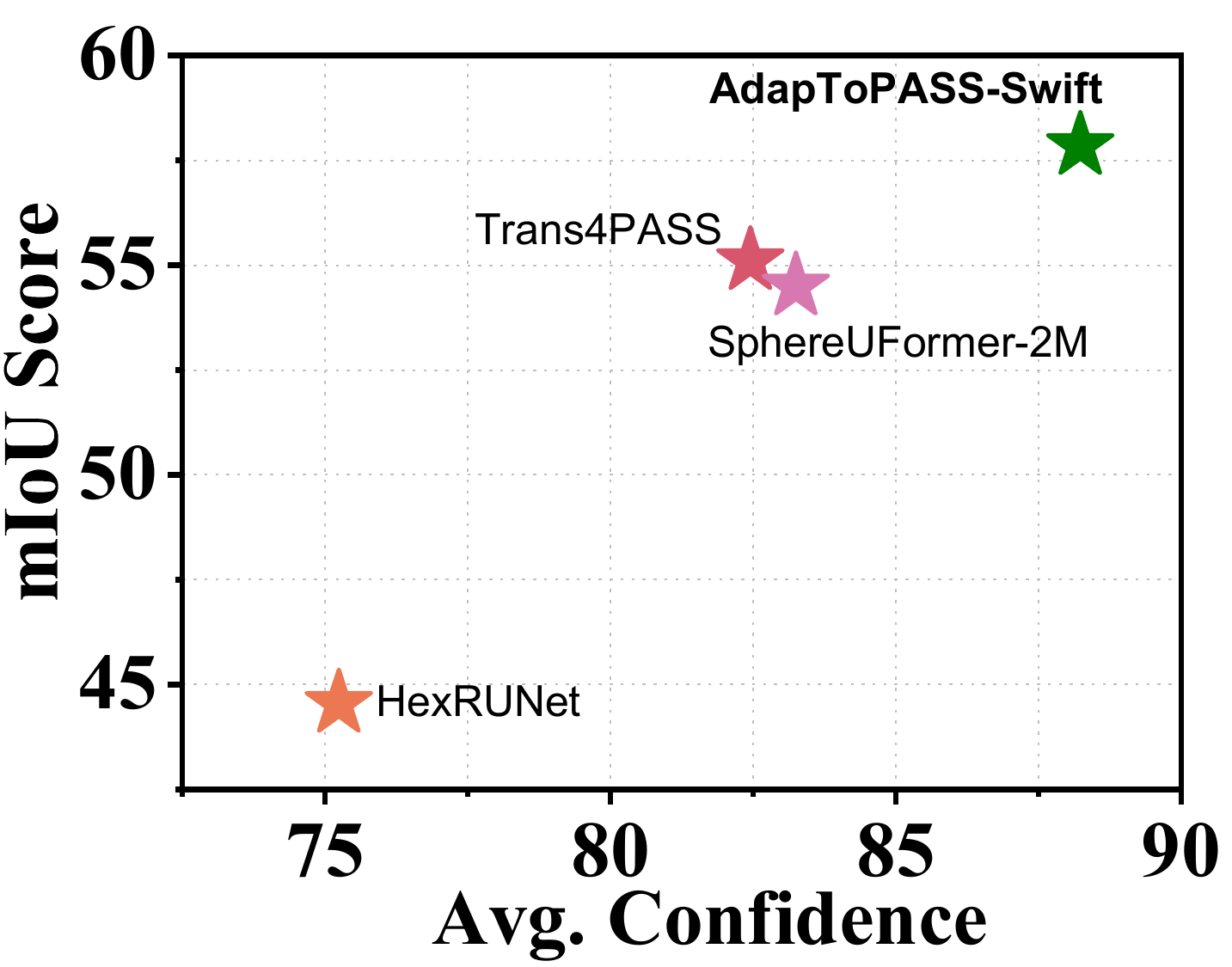}
        \vspace{-15pt}
        \caption{}
        \label{fig:quant_swift}
    \end{subfigure}
    \vspace{-5pt}
    \caption{\textbf{(a)} Analysis on Stanford2D3D under unseen spherical transformations.
        {(i)} Difficulty-wise \underline{mIoU} comparison across difficulties. 
         {(ii)} Transformation-wise \underline{mIoU} comparison under Rotation, Scale, Orientation Shift (Ori.), Translation, and Viewpoint Shift (View). \textbf{(b)} Comparison of AdapToPASS-Swift (our 2M param variant) with lightweight panoramic segmentation baselines.}
\vspace{-15pt}
\end{figure*}

We structure our evaluation around the following experimental questions (EQs):

\begin{center}
\vspace{-3pt}
\colorbox{orange!10}{
\begin{minipage}{0.96\linewidth}
\textbf{EQ-1:} Does modeling contextual and geometric ambiguity within a Spherical Transformer framework improve the accuracy and consistency of panoramic semantic segmentation?\vspace{2pt}\\
\textbf{EQ-2:} Does the bio-inspired ambiguity-aware design of AdapToPASS enable more reliable generalization than canonical methods on unseen spherical transformations?\vspace{2pt}\\
\textbf{EQ-3:} What is the impact of AdaSpA, Bifocal Spherical Representation, and boundary supervision on the overall performance of AdapToPASS?
\end{minipage}
}
\vspace{-3pt}
\end{center}

\begin{figure*}[t!]
    \centering
    \includegraphics[width=0.88\textwidth]{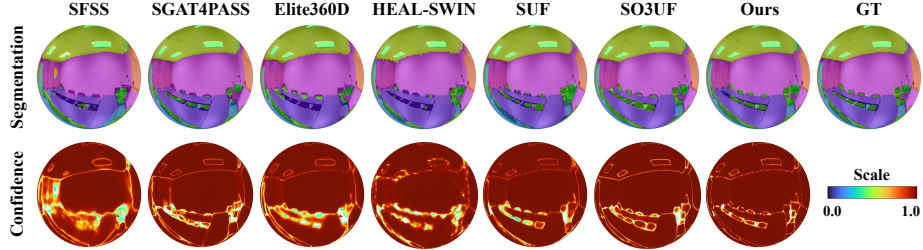}
      \vspace{-5pt}
    \caption{\textbf{Qualitative comparison on Stanford2D3D.}
We compare segmentation predictions and their corresponding confidence maps across representative panoramic segmentation methods. 
AdapToPASS produces more accurate semantic regions with sharper boundaries while maintaining high confidence (SUF: SphereUFormer, SO3UF: SO3UFormer). \textit{Zoom in for a better view}.}
    \label{fig:qualitative_stanford2d3d}
    \vspace{-15pt}
\end{figure*}

\noindent \textbf{Comparison with SOTA Methods.}
Table~\ref{tab:sota_comparison} and Fig~\ref{fig:teaser}\textcolor{DeepPink}{(b-c)} compares AdapToPASS against existing panoramic semantic segmentation methods on Stanford2D3D and WildPASS. We report the \underline{test set} mean Intersection-over-Union (mIoU) and mean class accuracy (mAcc)\footnote{mAcc denotes the average of per-class pixel accuracies with equal weighting across semantic categories, thereby reducing the dominance of large or frequent classes}. 
\vspace{-5pt}
\begin{center}
\vspace{-3pt}
\colorbox{gray!10}{
\begin{minipage}{0.96\linewidth}
    \textcolor{DeepPink}{\textbf{[A-1]}} 
AdapToPASS achieves the best overall performance across both datasets, obtaining \textbf{70.80\%} mIoU on Stanford2D3D, and \textbf{69.58\%} mIoU  on WildPASS. 
Notably, AdapToPASS produces the highest average prediction confidence (Conf.), indicating more reliable semantic predictions. 
\end{minipage}
}
\vspace{-3pt}
\end{center}
\vspace{-5pt}
Fig~\ref{fig:qualitative_stanford2d3d} illustrates the qualitative result on Stanford2D3D. Refer to Appendix~\ref{appendix:results} for additional results.

\noindent \textbf{Robustness Under Unseen Spherical Transformations.}
Tables~\ref{tab:robustness_results_stanford2d3d} and~\ref{tab:robustness_result_wildpass} evaluate robustness under unseen spherical transformations on Stanford2D3D and WildPASS. 
\vspace{-5pt}
\begin{center}
\vspace{-3pt}
\colorbox{gray!10}{
\begin{minipage}{0.96\linewidth}
    \textcolor{DeepPink}{\textbf{[A-2]}} 
AdapToPASS achieves the highest mIoU across all transformation types and severity levels. 
Figures~\ref{fig:analysis_stanford2d3d} and~\ref{fig:robustness_summary_wildpass} show that AdapToPASS achieves upto \textbf{+13.38\%} and \textbf{+18.77\%} \textbf{relative improvements in mIoU} on Stanford2D3D and WildPASS, respectively.
\end{minipage}
}
\vspace{-3pt}
\end{center}

While the robust and stress settings introduce severe geometric distortions, orientation shifts, and viewpoint changes that degrade all methods, AdapToPASS consistently retains a clear margin over prior approaches. 
These results demonstrate that bio-inspired, ambiguity-aware adaptive spherical perception enables stronger generalization to unseen spherical transformations, making AdapToPASS \textbf{closer to human perceptual behavior}, even without transformation-specific augmentation.

\begin{wraptable}{r}{0.48\textwidth}
    \centering
    \resizebox{0.48\textwidth}{!}{%
    \begin{tabular}{l|cc}
        \toprule
        \textbf{Configuration} & \textbf{mIoU\,$\uparrow$} & \textbf{Conf.\,$\uparrow$} \\
        \midrule
        AdapToPASS & 70.80 & 96.42 \\
        AdapToPASS w/o Context Ambiguity & 64.97 & 88.47 \\
        AdapToPASS w/o Acuity Stream & 66.50 & 92.20 \\
        AdapToPASS w/o Boundary Supervision & 63.70 & 91.90 \\
        \bottomrule
    \end{tabular}%
    }
    \vspace{-7pt}
    \caption{Ablation study evaluating the contribution of key components in AdapToPASS.}
    \label{tab:ablation}
    \vspace{-10pt}
\end{wraptable}

\noindent \textbf{Ablation Study.}
Table~\ref{tab:ablation} presents an ablation analysis on the Stanford2D3D benchmark to evaluate the contribution of each component in AdapToPASS. We observe several interesting findings: \textbf{(I)} Removing the contextual ambiguity modeling significantly reduces the overall prediction confidence;  \textbf{(II)} Removing the SDF supervision causes a notable drop in mIoU; \textbf{(III)} Using only contextual ambiguity with boundary supervision results a compatibility better mIoU and confidence.

\begin{center}
\vspace{-5pt}
\colorbox{gray!10}{
\begin{minipage}{0.96\linewidth}
    \textcolor{DeepPink}{\textbf{[A-3]}} 
AdapToPASS achieves the strongest performance when contextual ambiguity modeling, Bifocal Spherical Representation, and boundary supervision are jointly integrated. Their combination enables adaptive context aggregation, geometry-aware multi-scale representation, and sharper boundary localization. Fig.~\ref{fig:qual_ablation} illustrates the effect of each of these component.
\end{minipage}
}
\end{center}

\vspace{-5pt}

\noindent \textbf{\colorbox{green!10}{AdapToPASS-Swift.}}
We further introduce AdapToPASS-Swift, a lightweight variant of AdapToPASS with \underline{fewer than 2M parameters}. 
It follows the exact bio-inspired, ambiguity-aware spherical perception principle as AdapToPASS, but replaces AdaSpA with lightweight \textbf{Adaptive Spherical Context Aggregation Module} (\textbf{AdaSpX}) for efficient spherical semantic understanding. 
Under a matched compact-model setting, AdapToPASS-Swift generalizes better than similarly sized panoramic segmentation baselines, achieving stronger and more confident overall performance (Fig.~\ref{fig:quant_swift}, Table~\ref{tab:lite_comparison}) and robustness to unseen spherical transformations (Table~\ref{tab:lite_robustness_results}). 
For implementation details and baseline setup please refer to Appendix~\ref{appendix:adaptopass_swift} and~\ref{apps:baseline_implementation}.

\begin{wrapfigure}{r}{0.35\columnwidth}
\vspace{-10pt}
\centering
\includegraphics[width=\linewidth]{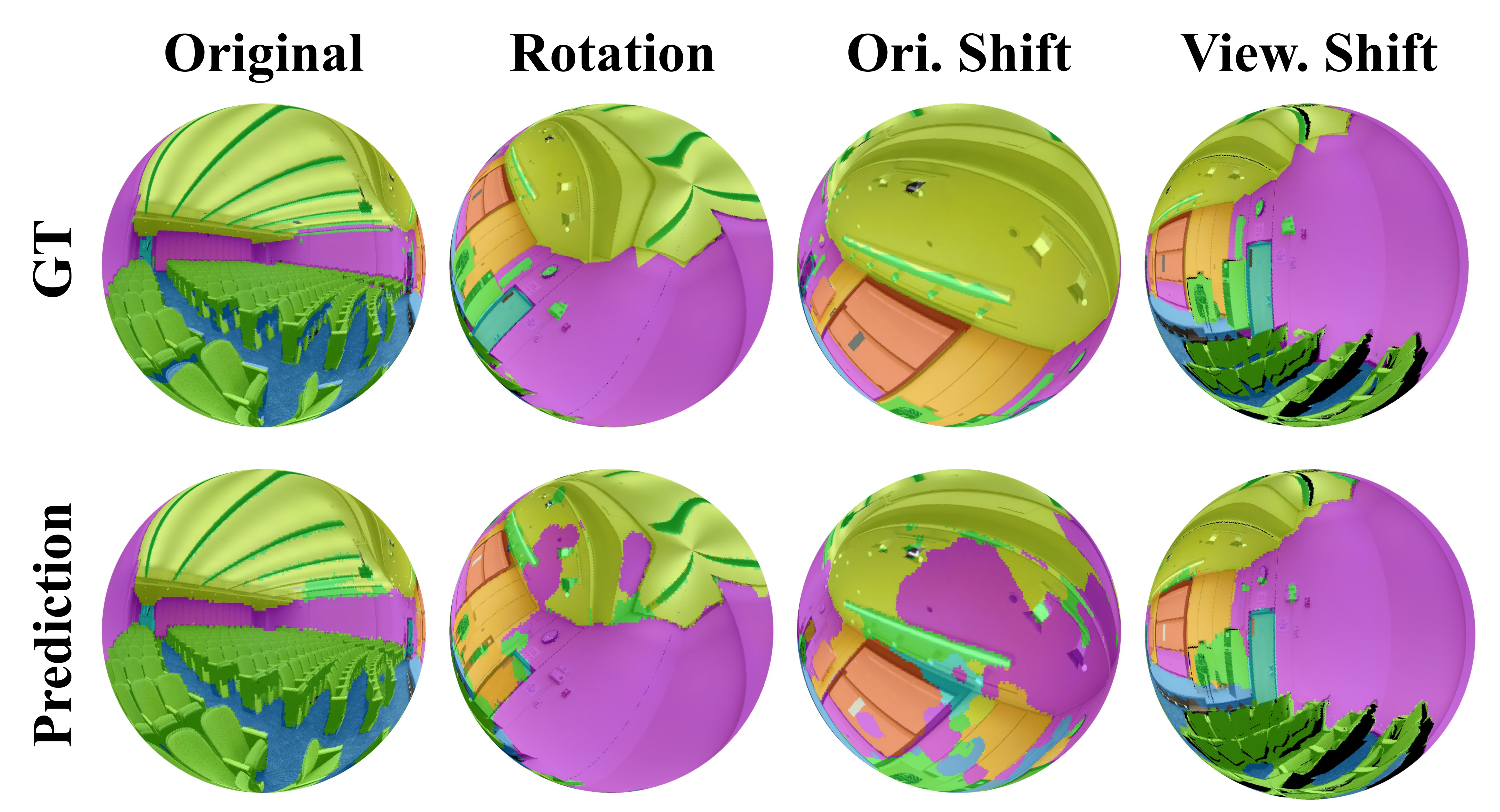}
\caption{Failure cases of AdapToPASS on Stanford2d3d under \textit{Stress} setting.}
\label{fig:fails}
\vspace{-15pt}
\end{wrapfigure}

\noindent\textbf{Failure Cases.} 
Although AdapToPASS consistently outperforms prior methods under unseen spherical transformations, the performance still degrades noticeably under extremely severe perturbations, especially in the stress setting (Fig~\ref{fig:analysis_stanford2d3d}). This degradation is most pronounced for \textit{orientation shifts} and \textit{viewpoint shifts}, where large changes in spherical layout can disrupt semantic consistency and introduce severe geometric ambiguity (Fig~\ref{fig:fails}). 

\vspace{-9pt}
\section{Conclusion and Future Work}
\vspace{-9pt}
In this work, we investigated the role of ambiguity-aware spherical perception for robust panoramic semantic segmentation. We introduced AdapToPASS, a bio-inspired Spherical Transformer framework inspired by uncertainty-driven contextual and geometric processing in biological vision. 
Comprehensive evaluations 
demonstrate that AdapToPASS consistently achieves strong segmentation performance. More importantly, it demonstrates significantly improved robustness to unseen spherical transformations compared to prior baselines. 

\textbf{Future Work:} We aim to extend AdapToPASS in two directions: scaling ambiguity-aware spherical perception toward foundation models, including integration with large segmentation models such as SAM3, and developing lightweight variants for efficient robotic deployment. We hope this work motivates further research on ambiguity-aware spherical perception for robust robotic vision.







\bibliographystyle{unsrt} 
\bibliography{references} 





\appendix
\section{Additional Methodological Details}
\label{appendix:methodology}

\subsection{Icosphere Representation}
\label{apps:icosphere_repp}

\paragraph{Icosphere construction.}
AdapToPASS discretises the unit sphere $\mathbb{S}^2$ using a subdivided
icosahedron (\emph{icosphere}).
Starting from the regular icosahedron ($V_0\!=\!12$ vertices, $F_0\!=\!20$
triangular faces), each subdivision step replaces every triangle with four
smaller triangles whose new mid-edge vertices are projected onto
$\mathbb{S}^2$.
After $r$ such steps, a rank-$r$ icosphere has
\begin{equation}
  V_r = 10\cdot 4^{r} + 2 \quad \text{vertices} \qquad \text{and} \qquad
  F_r = 20\cdot 4^r \quad \text{faces,}
  \label{eq:ico_counts}
\end{equation}
each subdivision increasing resolution by a factor of four.


\paragraph{ERP-to-icosphere sampling.}
Input panoramas are stored as equirectangular projections (ERP), where pixel
$(u, v)$ in an image of width $W$ and height $H$ corresponds to azimuth
$\theta\!=\!(u/W)\!\cdot\!360°-180°$ and polar angle
$\phi\!=\!(v/H)\!\cdot\!180°$.
To lift an ERP image onto the icosphere, we first convert each vertex normal
$\mathbf{p}_i\!=\!(x_i,y_i,z_i)$ to spherical coordinates:
\begin{equation}
  \phi_i = \arccos(z_i) \in [0^{\circ}, 180^{\circ}],
  \qquad
  \theta_i = \operatorname{atan2}(y_i,\, x_i) \in [-180^{\circ}, 180^{\circ}].
  \label{eq:spherical}
\end{equation}
These are mapped to the normalized grid coordinates expected by bilinear
sampling:
\begin{equation}
  \tilde{u}_i = \frac{\theta_i}{180},
  \qquad
  \tilde{v}_i = \frac{2\phi_i}{180} - 1,
  \label{eq:grid_coords}
\end{equation}
so that $(\tilde{u}_i, \tilde{v}_i) \in [-1,1]^2$ spans the full ERP canvas.
Each vertex then receives its feature vector by sampling the ERP image at
$(\tilde{u}_i, \tilde{v}_i)$ using bilinear interpolation for RGB and depth,
and nearest-neighbour interpolation for discrete semantic labels, with border
padding at the poles.
The result is a set of $V_r$ feature vectors $\{\mathbf{f}_i\}_{i=1}^{V_r}$
that serve as the input sequence to the AdapToPASS encoder.
At rank~$7$ (the default), this yields $163{,}842$ vertices, providing
sufficient angular resolution to resolve fine semantic boundaries while
remaining computationally tractable.

\begin{figure*}[h!]
    \centering
    \includegraphics[width=1.0\textwidth]{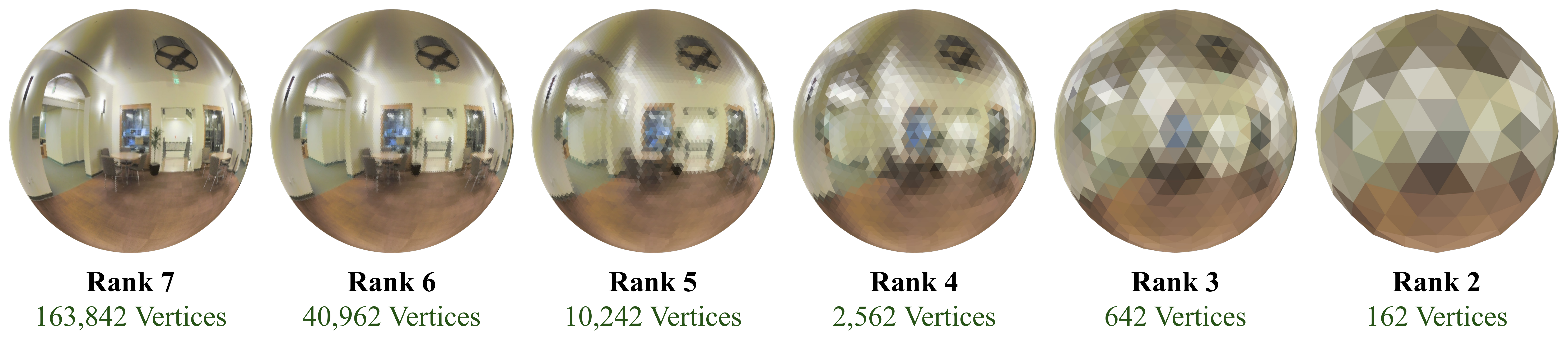}
    \caption{\textbf{Icosphere representations at different ranks.} The same panoramic scene is shown using different icosphere subdivision levels, ranging from Rank~2 (162 vertices) to Rank~7 (163,842 vertices). Higher ranks provide finer spherical resolution and preserve more visual detail.}
    \label{fig:ico_repp}
    \vspace{-10pt}
\end{figure*}

\paragraph{Hierarchical downsampling.}
A key structural property of the icosphere is that the vertex set at rank~$r{-}1$
is a strict subset of the vertex set at rank~$r$: every vertex of the coarser
mesh reappears verbatim after subdivision.
AdapToPASS exploits this to define a parameter-free \emph{center-select}
downsampling operator.
To move from rank~$r$ to rank~$r{-}1$, we precompute, for each of the
$V_{r-1}$ coarse vertices, the index of the matching fine vertex by exact
coordinate lookup, and then simply gather those entries:
$\mathbf{x}^{(r-1)} = \mathbf{x}^{(r)}[\,\mathcal{I}_{\downarrow}\,]$,
where $\mathcal{I}_{\downarrow} \in \mathbb{Z}^{V_{r-1}}$ is the precomputed
index map.
No learnable spatial parameters are introduced; the spatial information is
compressed purely by discarding the $V_r - V_{r-1}$ new vertices that were
added during the last subdivision.
Following each spatial downsampling step, a linear projection doubles the
channel dimension from $C$ to $2C$, so that the total feature capacity is
approximately preserved across scales.

\paragraph{Hierarchical upsampling.}
The decoder restores spatial resolution one rank at a time using a
geometry-aware \emph{interpolate} upsample operator.
When going from rank~$r$ to rank~$r{+}1$, the $V_r$ vertices that already
existed at the coarser rank inherit their features directly (identity copy).
The $V_{r+1} - V_r$ new mid-edge vertices, which are introduced by the
subdivision step and therefore have no direct coarse counterpart, receive
features by averaging the two coarse-rank edge-adjacent vertices:
\begin{equation}
  \mathbf{x}^{(r+1)}_i
  =
  \begin{cases}
    \mathbf{x}^{(r)}_i & i < V_r, \\[3pt]
    \dfrac{\mathbf{x}^{(r)}_{\ell(i)} + \mathbf{x}^{(r)}_{m(i)}}{2} & i \geq V_r,
  \end{cases}
  \label{eq:upsample}
\end{equation}
where $\ell(i)$ and $m(i)$ are the indices of the two rank-$r$ neighbors of
new vertex~$i$, determined once at construction time from the icosphere
adjacency.
This respects the subdivision geometry: new vertices lie at the midpoints of
coarse edges, so averaging the two endpoints is the exact linear interpolation
on the sphere.
Preceding each spatial upsampling step, a linear projection halves the channel
dimension from $2C$ to $C$.

\subsection{Training Objective and Optimization}
\label{apps:training_objective}

AdapToPASS is trained end-to-end with a unified objective combining semantic
segmentation supervision, SDF-based boundary terms, and an ambiguity
regularizer:
{\setlength\abovedisplayskip{3pt}
\setlength\belowdisplayskip{3pt}
\begin{equation}
\label{eq:obj}
\mathcal{L}_{\mathrm{overall}}
=
\mathcal{L}_{\mathrm{CE}}
+
\lambda_{\mathrm{Lov}}\,\mathcal{L}_{\mathrm{Lov}}
+
\lambda_{\mathrm{SDF}}\,\mathcal{L}_{\mathrm{SDF}}\!\left(\hat{\mathbf{S}},\,\mathbf{S}^{\mathrm{gt}}(\mathbf{D}^{\partial},\mathbf{Y})\right)
+
\lambda_{\mathrm{align}}\,\mathcal{L}_{\mathrm{align}}\!\left(\hat{\mathbf{Y}},\hat{\mathbf{S}}\right)
+
\lambda_{c}\,\bar{c}.
\end{equation}}
Here $\mathbf{Y}$ denotes the ground-truth semantic labels, $\hat{\mathbf{Y}}$
the predicted segmentation logits, $\mathbf{D}^{\partial}$ the unsigned
geodesic boundary-distance field on the icosphere graph, $\mathbf{S}^{\mathrm{gt}}(\mathbf{D}^{\partial},\mathbf{Y})$
the per-class signed distance field derived from $\mathbf{D}^{\partial}$ and
$\mathbf{Y}$, and $\hat{\mathbf{S}}$ the corresponding SDF predicted by a
dedicated head.
The SDF losses $\mathcal{L}_{\mathrm{SDF}}$ and $\mathcal{L}_{\mathrm{align}}$
are defined in \cref{eq:sdf_loss,eq:sdf_align} with $\lambda_{\mathrm{SDF}}\!=\!0.5$
and $\lambda_{\mathrm{align}}\!=\!0.1$.

The segmentation terms $\mathcal{L}_{\mathrm{CE}}$ and $\mathcal{L}_{\mathrm{Lov}}$
supervise the semantic head directly.
$\mathcal{L}_{\mathrm{CE}}$ is a class-balanced cross-entropy loss with
effective-number-of-samples class weights~\cite{cui2019class}, excluding the background class.
$\mathcal{L}_{\mathrm{Lov}}$ is the Lov\'{a}sz-Softmax
loss~\cite{berman2018lovasz}, a tight differentiable surrogate for the mean
intersection-over-union restricted to non-background classes, weighted by
$\lambda_{\mathrm{Lov}}\!=\!1.0$.

The final term $\lambda_{c}\,\bar{c}$ regularises the Contextual-Ambiguity-Aware Geodesic Bias (CxAGB) modules.
At each attention layer, a small MLP predicts a per-node ambiguity scalar
$u_i\!=\!\sigma(\mathrm{MLP}(\mathrm{LN}(\mathbf{x}_i)))\!\in\![0,1]$ that
continuously rescales the geodesic attention window:
$s_i\!=\!s_{\min}+(s_{\max}-s_{\min})\,u_i$, with $s_{\min}\!=\!0.8$ and
$s_{\max}\!=\!1.6$.
$\bar{c}$ is the mean of $u_i$ across all nodes and all CxAGB layers.

\paragraph{Optimization.}
AdapToPASS is trained with AdamW~\cite{zhou2024towards} using an
initial learning rate of $10^{-4}$, a minimum learning rate of $10^{-6}$,
weight decay of $0.01$, and a warmup-cosine schedule with 10 warmup epochs
over 400 total epochs.
Linear weights are initialised with a truncated normal distribution
($\sigma\!=\!0.02$); layer-normalisation parameters are initialised with
weight~$1$ and bias~$0$.
\textcolor{DeepPink}{\textbf{No~transformation-specific data augmentation is applied to AdapToPASS during training}}.


\section{Additional Experiment Details}

\label{appendix:experiments}

\subsection{Unseen Spherical Transformations}
\label{app:unseen_spherical_transformations}

We generate unseen transformed evaluation sets for Stanford2D3D and WildPASS to test robustness under spherical geometric perturbations. 
The transformation families are denoted as \textit{Rotation}, \textit{Scale}, \textit{Translation}, \textit{Orientation Shift}, and \textit{Viewpoint Shift}. 
\textit{Rotation} applies spherical yaw, pitch, and roll perturbations. 
\textit{Scale} is produced by applying a spherical M\"obius transformation, which conformally warps the sphere to simulate zoom while preserving local angular structure~\cite{zhou2021data}. 
We use a M\"obius zoom factor $s$, where $s>1$ concentrates content around the viewing direction to create zoom-in, and $s<1$ spreads content over a wider spherical region to create zoom-out. 
\textit{Translation} applies depth-based camera shifts along the 3D axes. 
\textit{Orientation Shift} combines spherical rotation with M\"obius zoom, while \textit{Viewpoint Shift} combines spherical rotation with depth-based translation.

For Stanford2D3D, we evaluate all five transformation families. 
For WildPASS, we evaluate \textit{Rotation}, \textit{Scale}, and \textit{Orientation Shift}; \textit{Translation} and \textit{Viewpoint Shift} are not used because depth-based viewpoint translation requires paired depth panoramas. 
For each transformation family, we define three difficulty regimes: \textit{mild}, \textit{robust}, and \textit{stress}. 
The \textit{mild} regime represents small perturbations close to nominal viewing conditions, \textit{robust} represents moderate viewpoint and scale changes, and \textit{stress} represents stronger geometric shifts for out-of-distribution robustness testing. The parameter candidates used for each transformation family and difficulty regime are summarized in Table~\ref{tab:difficulty_levels}.


\begin{table*}[h!]
\centering
\small
\setlength{\tabcolsep}{5pt}
\renewcommand{\arraystretch}{1.35}
\caption{\textbf{Difficulty levels for unseen spherical transformations.} Each level defines progressively stronger perturbations for rotation, scale, translation, orientation shift, and viewpoint shift.}
\vspace{5pt}
\label{tab:difficulty_levels}
\resizebox{\textwidth}{!}{%
\begin{tabular}{ll
>{\columncolor{green!8}}c
>{\columncolor{orange!5}}c
>{\columncolor{red!5}}c}
\toprule
\textbf{Type} 
& \textbf{Parameter}
& \cellcolor{green!15}\textbf{Mild} 
& \cellcolor{orange!15}\textbf{Robust} 
& \cellcolor{red!10}\textbf{Stress} \\
\midrule

\multirow{3}{*}{\textbf{Rotation}}
& \textbf{Yaw}
& $0^\circ,45^\circ,90^\circ,135^\circ,180^\circ$
& $0^\circ{:}30^\circ{:}330^\circ$
& $0^\circ{:}30^\circ{:}330^\circ$ \\

& \textbf{Pitch}
& $-10^\circ,-5^\circ,0^\circ,5^\circ,10^\circ$
& $-30^\circ,-20^\circ,-10^\circ,0^\circ,10^\circ,20^\circ,30^\circ$
& $-60^\circ,-45^\circ,-30^\circ,0^\circ,30^\circ,45^\circ,60^\circ$ \\

& \textbf{Roll}
& $-10^\circ,0^\circ,10^\circ$
& $-30^\circ,-15^\circ,0^\circ,15^\circ,30^\circ$
& $-45^\circ,0^\circ,45^\circ$ \\

\midrule

\textbf{Scale}
& \textbf{Zoom}
& $0.9,1.0,1.1,1.25,1.5$
& $0.8,1.0,1.2,1.5,2.0$
& $0.6,0.8,1.0,1.5,2.0,3.0$ \\

\midrule

\multirow{2}{*}{\textbf{Translation}}
& \textbf{$t_x/t_y$}
& $-0.08,-0.05,0,0.05,0.08$
& $-0.18,-0.12,-0.08,0,0.08,0.12,0.18$
& $-0.30,-0.22,-0.16,-0.10,0,0.10,0.16,0.22,0.30$ \\

& \textbf{$t_z$}
& $-0.04,0,0.04$
& $-0.08,-0.05,0,0.05,0.08$
& $-0.12,-0.08,0,0.08,0.12$ \\

\midrule

\textbf{Orientation Shift}
& \textbf{Composition}
& \multicolumn{3}{c}{Rotation + Scale} \\

\textbf{Viewpoint Shift}
& \textbf{Composition}
& \multicolumn{3}{c}{Rotation + Translation} \\

\bottomrule
\end{tabular}%
}
\end{table*}

\paragraph{Generation protocol.}
For each source panorama, we generate transformed evaluation samples independently for each difficulty regime and transformation family. 
The transformation parameters are sampled from predefined severity-specific candidate sets. 
For \textit{Rotation}, yaw, pitch, and roll are sampled from the corresponding angular ranges, while the scale factor is fixed to $1.0$. 
For \textit{Scale}, yaw, pitch, and roll are fixed to zero, and the M\"obius zoom factor is sampled from the corresponding zoom set. 
For \textit{Orientation Shift}, both spherical rotation and M\"obius zoom are applied jointly. 
Identity transformations are excluded: pure rotation and rotation-composed families do not allow zero yaw, pitch, and roll simultaneously, and pure scale and rotation-scale families do not use zoom factor $1.0$.

\begin{figure*}[h!]
    \centering
    \includegraphics[width=1.0\textwidth]{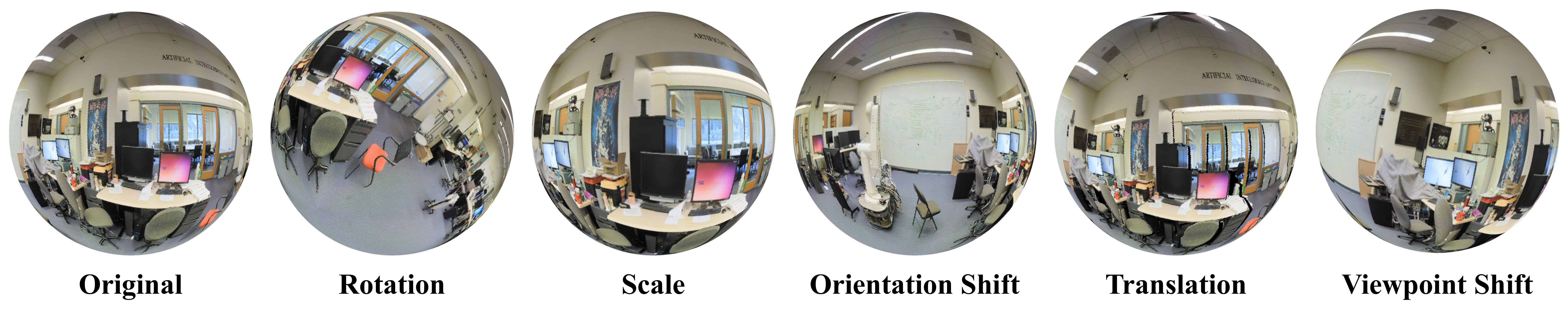}
    \vspace{-15pt}
    \caption{Examples of unseen spherical transformations.}
    \label{fig:transform_exmaple}
\end{figure*}

For translation-based transformations, we use the paired depth panorama to perform a geometric viewpoint shift rather than a purely image-space warp. 
Each valid ERP pixel is first lifted to a 3D point using its depth value and spherical ray direction. 
The camera is then translated by $(t_x,t_y,t_z)$, and for \textit{Viewpoint Shift}, the translated points are additionally rotated using the sampled yaw, pitch, and roll. 
The transformed 3D points are reprojected back to ERP coordinates, with collisions resolved using a nearest-depth rule. 
Pixels that receive no valid projection are left empty and recorded in an additional valid mask. 
Figure~\ref{fig:transform_exmaple} provides representative examples of the unseen spherical transformations used in our robustness evaluation.

\subsection{WildPASS Dataset and Label Space}
\label{app:wildpass_reduced}


The full WildPASS annotation set covers 8 Cityscapes categories~\cite{cordts2016cityscapes}; however,
the label distribution is highly skewed.
Three foreground classes, \emph{road}, \emph{sidewalk}, and \emph{car},
dominate the annotated pixel budget and together account for approximately
$92\%$ of all labelled pixels across the dataset.
The remaining categories each constitute a small minority and are frequently absent from
individual panoramas entirely.

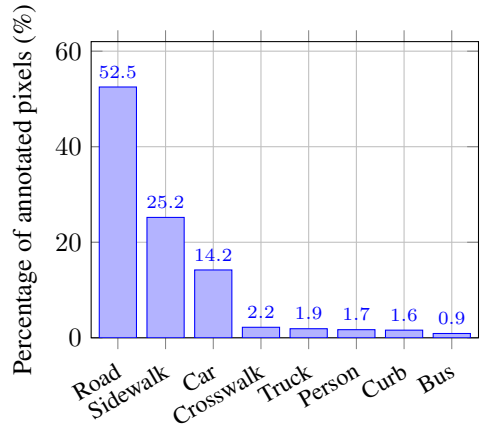
\begin{wrapfigure}{r}{0.5\textwidth}
    \centering
    \vspace{-2mm}
    \begin{tikzpicture}
        \begin{axis}[
            ybar,
            bar width=14pt,
            width=0.48\textwidth,
            height=5.5cm,
            ymin=0,
            ymax=62,
            ylabel={Percentage of annotated pixels (\%)},
            symbolic x coords={Road, Sidewalk, Car, Crosswalk, Truck, Person, Curb, Bus},
            xtick=data,
            xticklabel style={font=\small, rotate=30, anchor=north east},
            nodes near coords,
            nodes near coords style={font=\scriptsize},
            nodes near coords align={vertical},
            enlarge x limits=0.08,
            grid=major,
            every axis plot/.append style={
                fill=gray!30,
                draw=black,
            },
        ]
        \addplot coordinates {
            (Road,      52.5)
            (Sidewalk,  25.2)
            (Car,       14.2)
            (Crosswalk,  2.2)
            (Truck,      1.9)
            (Person,     1.7)
            (Curb,       1.6)
            (Bus,        0.9)
        };
        \draw [decorate, decoration={brace, amplitude=5pt, raise=18pt},
               thick, gray]
            (axis cs:Road,54) -- (axis cs:Car,54)
            node [midway, above=24pt, font=\scriptsize, align=center]
            {$91.8\%$ of annotated pixels};
        \end{axis}
    \end{tikzpicture}
    \vspace{-5pt}
    \caption{Class distribution for WildPASS dataset.}
    \label{fig:wildpass_hist}
\end{wrapfigure}

We therefore collapse all categories except road, sidewalk, and car into a
single \emph{background} class (label~$0$), which is excluded from loss
computation and evaluation.
This gives a 4-class label space $\{0, 1, 2, 3\}$ = \{background, road,
sidewalk, car\} used consistently across all experiments.
As shown in Fig.~\ref{fig:wildpass_hist}, the WildPASS annotations are highly imbalanced, with Road, Sidewalk, and Car accounting for the majority of foreground pixels.

Our primary goal on WildPASS is to evaluate robustness and transformation
invariance under spherical perturbations, not general scene parsing.
Restricting evaluation to the three dominant classes serves two purposes.
First, it ensures that metrics are not dominated by the large number of
background pixels or by classes with too few samples to yield reliable mIoU
estimates.
Second, road, sidewalk, and car exhibit qualitatively different geometric
footprints on the sphere,  a broad ground-level stripe, a narrow transitional
band, and compact mid-field objects, respectively, making them collectively
informative for assessing how well a model preserves spatially structured
predictions under viewpoint variation.
Evaluating all methods under this identical, fixed label space ensures a fair
and reproducible comparison.

\subsection{Baseline Implementation Details}
\label{apps:baseline_implementation}

We summarize the implementation details used for the baseline comparisons. 
Patch-based foundation-model approaches are excluded from the comparison due to their substantially larger parameter counts, computational cost, and different inference paradigm.

\begin{center}
\colorbox{orange!10}{
\begin{minipage}{0.96\linewidth}
    \textbf{Note.} 
    For a fair comparison, all models are trained from scratch with RGB inputs only, using comparable model settings when variants exist. Predictions from different representations are evaluated on a common uniformly sampled sphere to avoid ERP-induced bias. All models are trained bsaed on their suggested epoch
\end{minipage}
}
\end{center}

Whenever the original implementation supports multiple model sizes, we choose or configure the variant whose depth, number of downsampling stages, and parameter count are closest to our model. 
Specifically, we use four downsampling stages and a depth of two blocks per stage whenever supported by the architecture. 
For grid-based models, we evaluate input resolutions of $256\times512$ and $512\times1024$; for spherical models, we choose the closest icosahedral or spherical-grid resolution in terms of sample count. 
All predictions are projected uniformly onto the sphere before evaluation, avoiding the latitude bias introduced by direct ERP evaluation.

\paragraph{SFSS~\cite{guttikonda2024single}.}
SFSS is evaluated in its unimodal RGB setting so that no depth or normal modality is used. 
Although the original RGB-X dataloader retains an auxiliary tensor, the unimodal model uses a single-stream RGB backbone and ignores the auxiliary input. 
We use a SegFormer-style~\cite{xie2021segformer} encoder with a DMLP decoder~\cite{tu2023learning}, and enable deformable convolution only in the first stage. 
The model is trained from scratch using AdamW with learning rate $6\times10^{-5}$, weight decay $0.01$, polynomial learning-rate decay with power $0.9$, and 10 warm-up epochs. 
Input panoramas are resized to $512\times1024$. 
Training uses ImageNet normalization, random horizontal flipping, multi-scale augmentation, and random $512\times512$ crops. 
Evaluation uses single-scale sliding-window inference without flip augmentation.

\paragraph{SGAT4PASS~\cite{li2023sgat4pass}.}
SGAT4PASS is reproduced using its Trans4PASS-style encoder-decoder architecture under the same RGB-only setting. 
The encoder follows a MiT-style hierarchy with deformable convolution enabled only in the first stage and dense decoder enhancement enabled across all stages~\cite{xie2021segformer}. 
All weights are randomly initialized. 
Training uses cross-entropy loss with AdamW and learning rate $8\times10^{-5}$. 
The model is trained with square crops and reprojection augmentation, including rotations along the $x$, $y$, and $z$ axes. 
Evaluation follows the spherical evaluation protocol described above after projecting predictions to the sphere.

\paragraph{HEAL-SWIN~\cite{carlsson2024heal}.}
HEAL-SWIN is evaluated as a spherical-grid Transformer based on the HEALPix representation. 
The model uses a Swin-style hierarchy with patch-based tokenization, shifted-window attention, and flat relative-position bias. 
We use the closest model scale to the other baselines, with four stages and hierarchical attention blocks. 
The model is trained from scratch with mixed precision using learning rate $10^{-4}$ and weight decay $10^{-4}$. 
Since HEAL-SWIN operates natively on a spherical grid, its output is projected to the common spherical evaluation grid for fair comparison.

\paragraph{Elite360D~\cite{ai2024elite360d}.}
Elite360D is evaluated using its ERP-plus-icosahedral design. 
The model combines an ERP branch with an icosahedral branch using a ResNet-18 backbone and face-level icosahedral features. 
The spherical branch uses an icosphere level chosen to match the pixel/sample count of the grid-based baselines as closely as possible. 
The model is trained from scratch using Adam with learning rate $10^{-4}$ and AMP enabled. 
Color augmentation, horizontal flipping, and yaw augmentation are enabled during training (matching the original setup). 
Final predictions are projected to the sphere for uniform evaluation.

\paragraph{SphereUFormer~\cite{benny2025sphereuformer}.}
SphereUFormer is evaluated as a vertex-icosphere Transformer operating directly on the spherical domain. 
We use a rank-7 icosphere with four hierarchical scales, depth two per scale, encoder heads $[2,4,8,16]$, and decoder heads $[16,16,8,4]$. 
The model uses center downsampling, interpolation-based upsampling, and both absolute and relative positional encodings. 
Training is performed from scratch using Adam with learning rate $10^{-4}$. 
Color augmentation is disabled, while horizontal flipping and yaw rotation remain enabled (matching the original setup). 
Since the model already predicts on the icosphere, its output is evaluated on the common spherical grid.

\paragraph{SO3UFormer~\cite{zhu2026so3uformer}.}
SO3UFormer follows the same hierarchical icosphere structure as SphereUFormer but replaces standard spherical attention and resampling with SO(3)-aware components. 
It uses quadrature attention, gauge-aware relative-position bias, pool-invariant gauge mode, multiple gauge frames, area-average downsampling, geodesic-kernel upsampling, and an equivariance regularization term with weight $0.05$. 
The model is trained from scratch in \texttt{bf16} precision using Adam with learning rate $10^{-4}$. 
Color augmentation is disabled, while horizontal flipping and yaw rotation remain enabled (matching the original setup). 
Predictions are evaluated on the same uniformly sampled spherical surface as the other methods.

\paragraph{HexRUNet~\cite{zhang2019orientation}.}
HexRUNet is implemented as a lightweight spherical encoder-decoder operating on an icosahedral level-4 representation. 
The model follows a compact residual U-Net design with two downsampling blocks, two decoder blocks, and a final per-face segmentation classifier. 
Starting from a base width of 93 channels, the encoder progressively expands the feature dimension to 186 and 372 channels before decoding back to the semantic output space. 
We train HexRUNet from scratch with AdamW, using learning rate $10^{-4}$, weight decay $10^{-4}$, and a StepLR schedule with decay factor $0.5$. 
Training is performed with mixed precision and batch size 64. 
For this baseline, yaw-rotation and horizontal-flip augmentations are disabled.

\paragraph{Trans4PASS~\cite{zhang2022bending}.}
Trans4PASS is implemented using the compact micro configuration for panoramic semantic segmentation. 
The model follows a lightweight encoder-decoder design with embedding dimension 60 and backbone channel widths $[32,64,128,192]$. 
Deformable convolution is used only in the first stage, while dense decoder enhancement is applied at all stages. 
We train Trans4PASS from scratch without external pretrained initialization. 
Optimization uses AdamW with learning rate $5\times10^{-5}$ and cross-entropy loss. 
The model is trained using $1080\times1080$ crops and evaluated using high-resolution $2048\times1024$ crops following the panoramic inference protocol.

\paragraph{SphereUFormer-2M~\cite{benny2025sphereuformer}.}
SphereUFormer-2M is implemented as a compact spherical Transformer operating directly on a rank-7 vertex icosphere. 
The model follows a three-scale hierarchical encoder-decoder design with embedding dimension 27, scale depth 2, encoder heads $[1,3,6]$, and decoder heads $[6,3,1]$. 
The network uses center downsampling, interpolation-based upsampling, and local spherical attention with checkpointing. 
We train SphereUFormer-2M from scratch using Adam with learning rate $10^{-4}$, batch size 32, and drop-path rate $0.05$. 
Following the full SphereUFormer setting, color augmentation is disabled, while horizontal flipping and yaw rotation remain enabled.


\subsection{Hyperparameter Details for AdapToPASS}
\label{apps:hyperparam_adaptopass}

AdapToPASS operates on rank-7 icosphere vertices with RGB inputs and an ERP resolution of $512 \times 256$. The base embedding dimension is 32, with four encoder-decoder scales, depth 2 per encoder/decoder scale, bottleneck depth 2, encoder heads $[2,4,8,16]$, bottleneck heads 16, and decoder heads $[16,16,8,4]$. The Acuity Stream uses depth 1, 2 heads, and a dimension ratio of 2. 

We train AdapToPASS using AdamW for 300 epochs with batch size 4 per GPU, learning rate $10^{-4}$, minimum learning rate $10^{-6}$, weight decay 0.01, and a warmup-cosine schedule with 10 warmup epochs. The drop-path rate is 0.05, and no explicit data augmentation is used. RGB inputs are normalized with mean 0.5 and standard deviation 0.225. 
\section{Additional Results and Discussions}
\label{appendix:results}

\begin{figure*}[h!]
    \centering
    \includegraphics[width=0.9\textwidth]{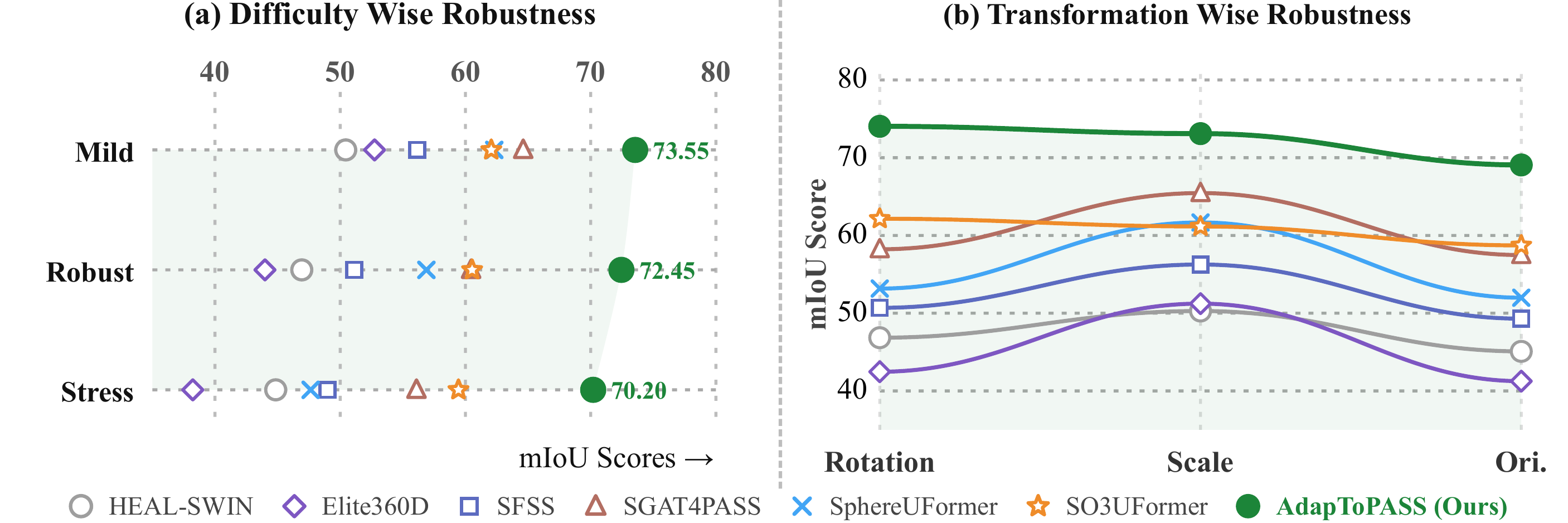}
    \caption{\textbf{Robustness analysis on WildPASS under unseen spherical transformations.}
\textbf{(a)} Difficulty-wise average mIoU comparison across mild, robust, and stress settings.
\textbf{(b)} Transformation-wise average mIoU comparison across rotation, scale, and orientation shift (Ori.).}
    \label{fig:robustness_summary_wildpass}
\end{figure*}

\begin{figure*}[h!]
    \centering
    \includegraphics[width=1.0\textwidth]{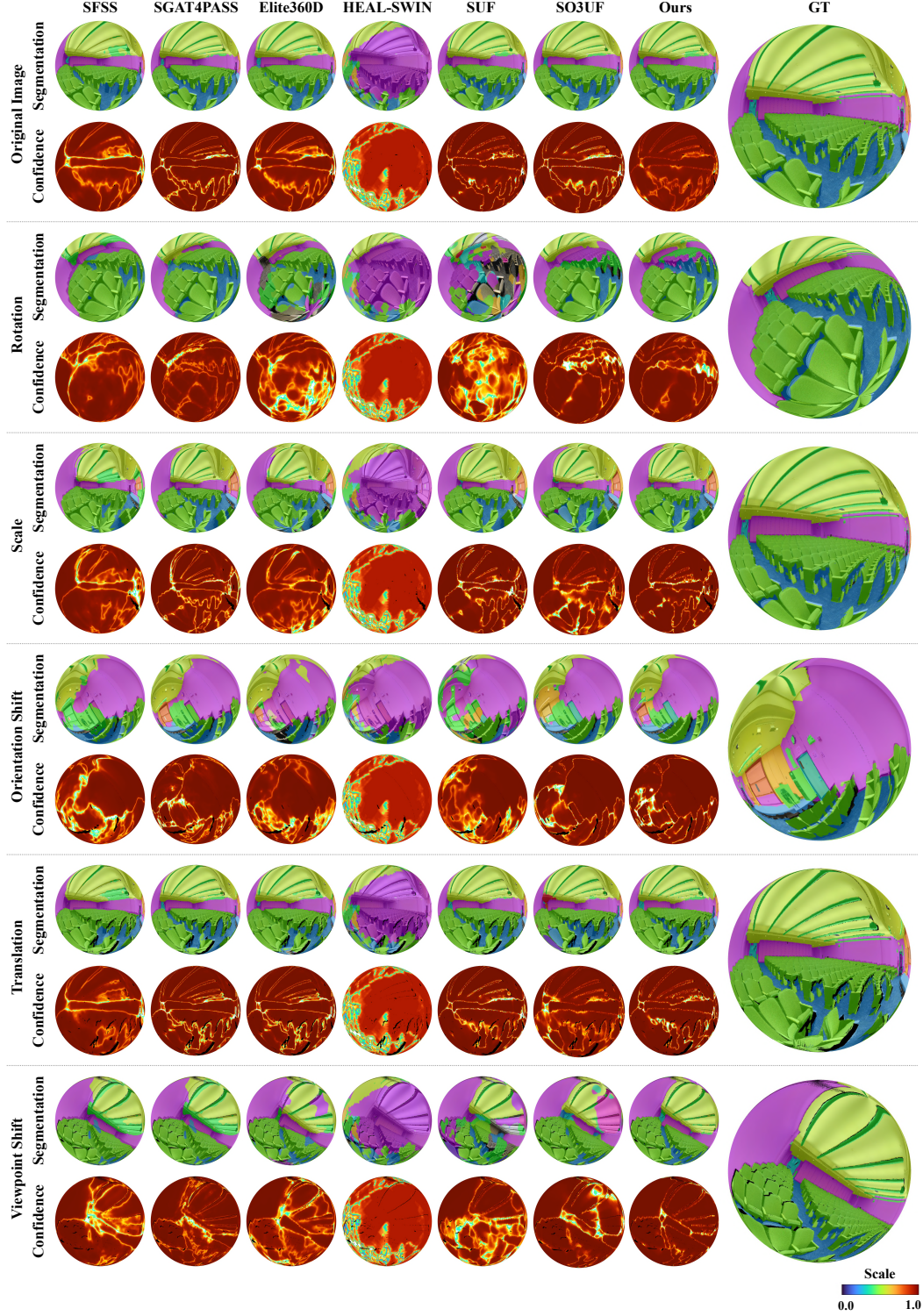}
    \caption{\textbf{Qualitative comparison on Stanford2D3D under unseen spherical transformations.}
We compare segmentation predictions and confidence maps for different methods across the original image and transformed variants, including rotation, scale, orientation shift, translation, and viewpoint shift under Robust setting.
Each row pair shows the predicted semantic map and the corresponding confidence map for the same input. \textit{Zoom in for a better view.}  (SUF:SphereUFormer, SO3F: SO3UFormer)}
\label{fig:wildpass_qualitative}
    \label{fig:full_stanford2d3d}
\end{figure*}

\begin{figure*}[h!]
    \centering
    \includegraphics[width=1.0\textwidth]{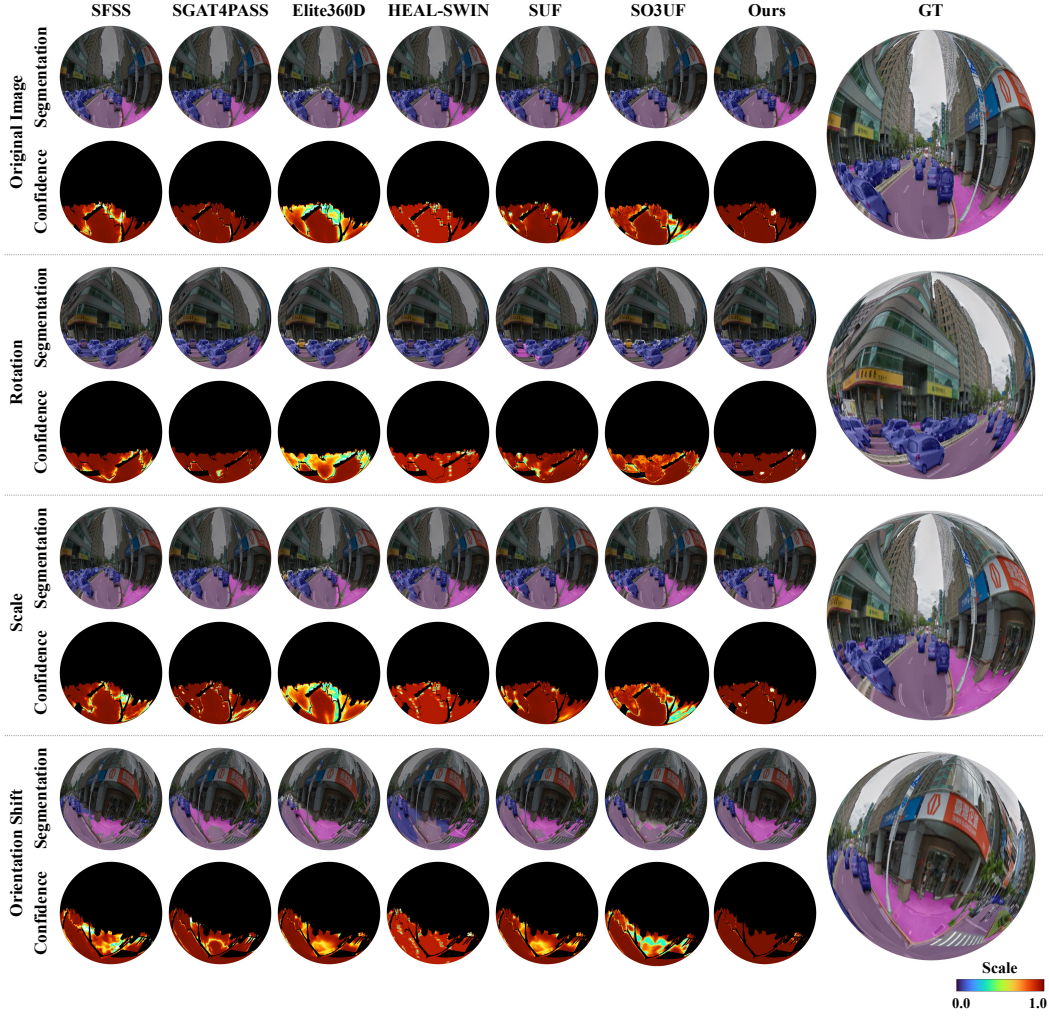}
    \caption{\textbf{Qualitative comparison on WildPASS under unseen spherical transformations.}
We compare segmentation predictions and confidence maps for different methods across the original image and transformed variants, including rotation, scale, and orientation shift under Robust setting.
Each row pair shows the predicted semantic map and the corresponding confidence map for the same input. \textit{Zoom in for a better view.} (SUF:SphereUFormer, SO3F: SO3UFormer)}
\label{fig:wildpass_qualitative}
    \label{fig:full_wildpass}
\end{figure*}

\begin{enumerate}
    \item Table~\ref{tab:robustness_result_wildpass} reports the mIoU comparison on WildPASS under unseen spherical transformations across mild, robust, and stress settings. 
    AdapToPASS consistently outperforms all baselines across rotation, scale, and orientation shifts, with the gains becoming especially clear under the more challenging robust and stress settings.

    \item Fig.~\ref{fig:robustness_summary_wildpass} further summarizes the WildPASS robustness results by averaging performance across difficulty levels and transformation families.
    AdapToPASS consistently achieves the highest mIoU across both difficulty levels and transformation types, showing stronger robustness under challenging spherical perturbations.

    \item Fig.~\ref{fig:full_stanford2d3d} and Fig.~\ref{fig:full_wildpass} provides a qualitative comparison on Stanford2D3D and WildPASS datasets respectively under unseen spherical transformations.
    While existing methods often show fragmented predictions and degraded confidence after geometric perturbations, AdapToPASS maintains more consistent semantic structure across transformations.
    This suggests that ambiguity-aware spherical modeling improves both prediction stability and confidence under challenging panoramic viewpoint changes.

    \item \textbf{Social Impact.} AdapToPASS can support safer and more reliable panoramic perception for embodied systems such as mobile robots, assistive agents, and immersive AR/VR platforms. By improving robustness under camera motion and unseen spherical transformations, it may reduce perception failures in real-world 360$^\circ$ environments where stable viewpoint assumptions do not hold. Its lightweight variant further suggests potential for deployment on resource-constrained robotic systems. However, as with any scene understanding model, incorrect segmentation may still lead to downstream errors in safety-critical applications; therefore, deployment should include uncertainty monitoring, human oversight, and careful validation across diverse environments.

    
\end{enumerate}

\begin{table}[h!]
\begin{center}
    
\resizebox{0.8\textwidth}{!}{%
\begin{tabular}{l 
>{\columncolor{green!6}}c >{\columncolor{green!6}}c >{\columncolor{green!6}}c
>{\columncolor{orange!8}}c >{\columncolor{orange!8}}c >{\columncolor{orange!8}}c
>{\columncolor{red!6}}c >{\columncolor{red!6}}c >{\columncolor{red!6}}c}
\toprule
\multirow{2}{*}{\textbf{Method}} 
& \multicolumn{3}{c}{\cellcolor{green!15}\textbf{Mild}} 
& \multicolumn{3}{c}{\cellcolor{orange!18}\textbf{Robust}} 
& \multicolumn{3}{c}{\cellcolor{red!15}\textbf{Stress}} \\
\cmidrule(lr){2-4} \cmidrule(lr){5-7} \cmidrule(lr){8-10}
& \textbf{Rot.} & \textbf{Scale} & \textbf{Ori.}
& \textbf{Rot.} & \textbf{Scale} & \textbf{Ori.}
& \textbf{Rot.} & \textbf{Scale} & \textbf{Ori.} \\
\midrule
SFSS               & 56.53 & 58.43 & 53.46 & 49.21 & 56.58 & 47.59 & 46.33 & 53.78 & 46.85 \\
SGAT4PASS          & 64.32 & 66.60 & 62.91 & 58.80 & 65.88 & 56.67 & 51.53 & 63.88 & 52.87 \\
\midrule
Elite360D          & 53.14 & 55.09 & 50.00 & 41.26 & 51.17 & 39.55 & 33.02 & 47.44 & 34.23 \\
HEAL-SWIN          & 50.56 & 52.52 & 48.23 & 46.33 & 50.04 & 44.41 & 43.65 & 48.27 & 42.63 \\
SphereUFormer      & 61.75 & 63.88 & 61.28 & 55.04 & 62.15 & 53.46 & 42.72 & 58.98 & 41.22 \\
SO3UFormer         & 62.66 & 62.17 & 61.36 & 61.98 & 61.07 & 58.53 & 61.82 & 60.23 & 56.28 \\
\midrule
\rowcolor{gray!12}
\textbf{AdapToPASS (Ours)}  
& \textbf{74.09} & \textbf{75.05} & \textbf{71.52}
& \textbf{74.29} & \textbf{73.09} & \textbf{69.98}
& \textbf{73.74} & \textbf{71.15} & \textbf{65.70} \\
\bottomrule
\end{tabular}%
}
\vspace{5pt}
\caption{\textbf{mIoU comparison under unseen spherical transformations on the WildPASS benchmark.} Results are reported across mild, robust, and stress settings, including rotation (Rot.), Scale, and Orientation Shifts (Ori.). Higher is better.}
\label{tab:robustness_result_wildpass}
\end{center}
\end{table}

\begin{figure*}[t!]
    \centering
    \includegraphics[width=1.0\textwidth]{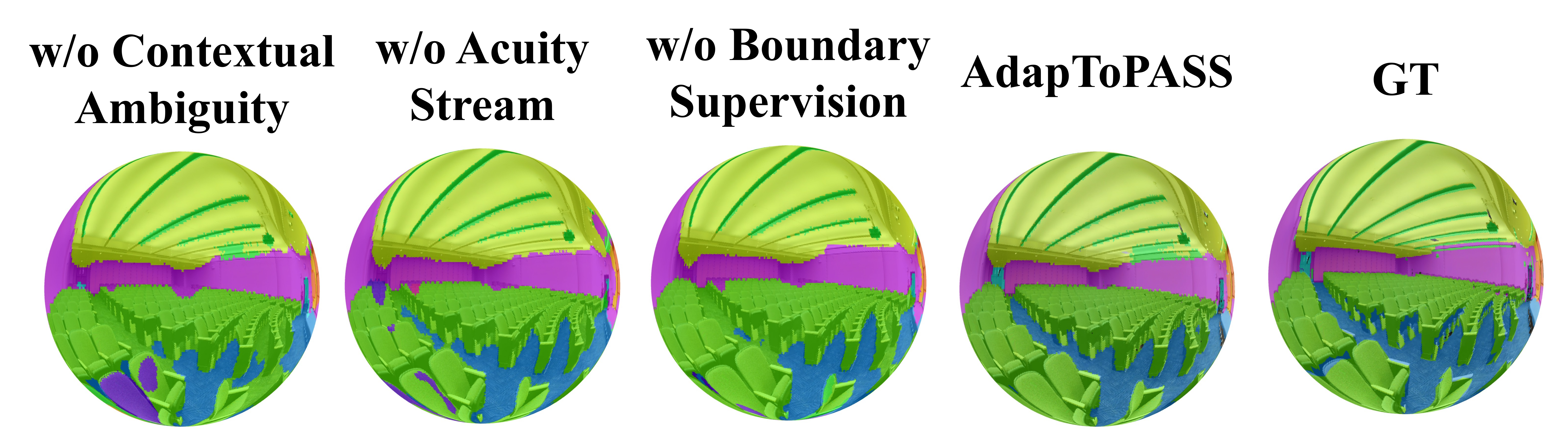}
    \caption{\textbf{Qualitative ablation study of AdapToPASS.}
Removing contextual ambiguity modeling, the acuity stream, or boundary supervision degrades semantic consistency and boundary quality. The full AdapToPASS model produces more coherent predictions that better align with the ground truth.}
    \label{fig:qual_ablation}
\end{figure*}

\section{AdapToPASS-Swift}
\label{appendix:adaptopass_swift}

\paragraph{Overview.}

\begin{wraptable}{r}{0.52\textwidth}
    \vspace{-10pt}
    \centering
    \resizebox{0.52\textwidth}{!}{%
    \begin{tabular}{l|c|ccc}
        \toprule
        \multirow{2}{*}{\textbf{Model}} 
        & \textbf{Params} 
        & \multicolumn{3}{c}{\textbf{Stanford2D3D}} \\
        
        & \textbf{(M)} 
        & \textbf{Acc.\,$\uparrow$} 
        & \textbf{mIoU\,$\uparrow$} 
        & \textbf{Conf.\,$\uparrow$} \\
        
        \midrule
        HexRUNet         & 1.99 & 59.79 & 44.55 & 75.24 \\
        SphereUFormer    & 2.04 & 69.42 & 55.10 & 82.45 \\
        Trans4PASS       & 1.99 & 70.60 & 54.50 & 83.25 \\
        
        \midrule
        \rowcolor{gray!10}
        \textbf{AdapToPASS-Swift}
        & 1.99
        & \textbf{73.20}
        & \textbf{57.85}
        & \textbf{88.23} \\
        
        \bottomrule
    \end{tabular}%
    }
    \caption{\textbf{Comparison of lightweight panoramic semantic segmentation models on Stanford2D3D.} We report average class accuracy (Acc.), mean IoU (mIoU), and mean prediction confidence (Conf.)}
    \label{tab:lite_comparison}
\end{wraptable}

To further examine the efficiency and robustness tradeoff in panoramic semantic segmentation, we introduce \textbf{AdapToPASS-Swift}, a lightweight variant of AdapToPASS with \textbf{less than 2M~params}, designed for resource constrained spherical perception. The goal of AdapToPASS-Swift is to preserve the key principles of ambiguity aware spherical modeling, including geometry aware local aggregation, adaptive contextual integration, and multi scale spherical perception, while reducing architectural complexity and computational overhead.

AdapToPASS-Swift follows the same high level design philosophy as AdapToPASS. It operates directly on the icosphere representation and maintains two complementary streams: an \textit{Acuity stream} for preserving fine local structure and a \textit{Lateral stream} for capturing broader contextual cues. Instead of relying on attention based token aggregation, AdapToPASS-Swift uses lightweight geometry conditioned operations to aggregate local and surround information on the sphere.

\subsection{AdaSpX: Adaptive Spherical Context Aggregation}
\label{app:adaspx}

\begin{figure*}[h!]
    \centering
    \includegraphics[width=0.95\textwidth]{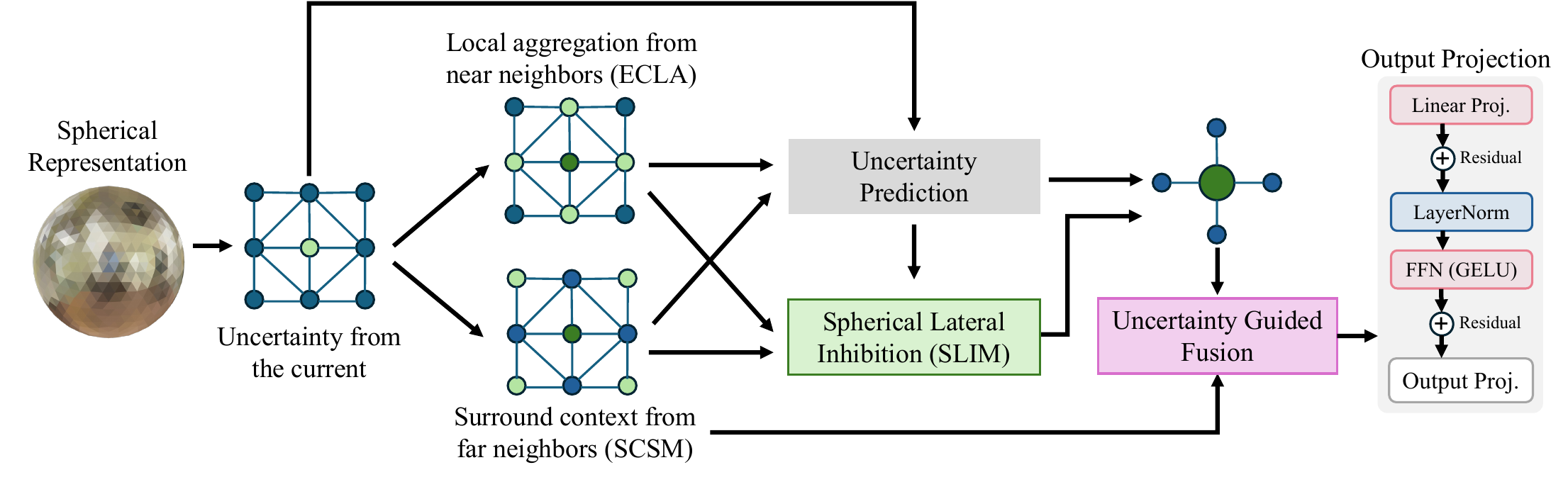}
    \caption{\textbf{Overview of the AdaSpX Module.} AdaSpX replaces the AdaSpA blocks in AdapToPASS to enable efficient spherical semantic segmentation.}
    \label{fig:methodology_adaptopass_swift}
    \vspace{-10pt}
\end{figure*}

At the core of AdapToPASS-Swift is the \textbf{Adaptive Spherical Context Aggregation} (\textbf{AdaSpX}) module (illustrated in Fig.~\ref{fig:methodology_adaptopass_swift}), an attention free spherical aggregation block for efficient ambiguity aware perception on the icosphere. Given node features $\mathbf{x}\in\mathbb{R}^{B\times N\times C}$ on the icosphere graph, AdaSpX aggregates information from two precomputed geodesic neighborhoods: a near neighborhood $\mathcal{N}_{\mathrm{near}}(i)$ for local geometric evidence and a far neighborhood $\mathcal{N}_{\mathrm{far}}(i)$ for wider spherical context.

\paragraph{Edge Conditioned Local Aggregation (ECLA).}
For each node $v_i$ and neighbor $v_j\in\mathcal{N}_{\mathrm{near}}(i)$, we construct an edge descriptor
\begin{equation}
\mathbf{e}_{ij}
=
[\mathbf{x}_i;\mathbf{x}_j;\mathbf{x}_j-\mathbf{x}_i;d_{ij}],
\end{equation}
where $d_{ij}$ is the geodesic distance between the corresponding icosphere vertices. A lightweight MLP predicts a channel wise gate
\begin{equation}
\mathbf{g}_{ij}
=
\sigma(\mathrm{MLP}(\mathbf{e}_{ij})),
\end{equation}
which modulates the neighbor message:
\begin{equation}
\mathbf{m}_{ij}
=
\mathbf{g}_{ij}
\odot
\frac{1}{d_{ij}+\epsilon}
\mathbf{W}_{v}\mathbf{x}_{j}.
\end{equation}
The local feature $\mathbf{l}_i$ is obtained by gated aggregation over the near neighborhood, together with a residual center projection. This provides a lightweight geometry aware mechanism for local spherical reasoning.

\paragraph{Surround Context Stabilization Module (SCSM).}
To incorporate broader contextual evidence, AdaSpX also aggregates from the far neighborhood $\mathcal{N}_{\mathrm{far}}(i)$. A scalar edge gate is predicted for each far neighbor and used to compute a distance weighted surround representation:
\begin{equation}
\mathbf{s}_i
=
\mathrm{SCSM}(\mathbf{x}_i,\{\mathbf{x}_j,d_{ij}\}_{j\in\mathcal{N}_{\mathrm{far}}(i)}).
\end{equation}
This term provides each node with a stable contextual reference, allowing ambiguous local regions to be interpreted with support from a wider spherical field of view.

\paragraph{Spherical Lateral Inhibition Module (SLIM).}
AdaSpX further applies a lightweight inhibition mechanism to suppress unstable local responses. Given the local feature $\mathbf{l}_i$, we compute a neighborhood consensus
\begin{equation}
\bar{\mathbf{l}}_i
=
\frac{1}{|\mathcal{N}_{\mathrm{near}}(i)|}
\sum_{j\in\mathcal{N}_{\mathrm{near}}(i)}
\mathbf{l}_j .
\end{equation}
The inhibition strength is predicted as
\begin{equation}
\boldsymbol{\alpha}_i
=
\sigma\left(
\mathrm{MLP}
([\mathbf{l}_i;\bar{\mathbf{l}}_i;\mathbf{l}_i-\bar{\mathbf{l}}_i;u_i])
\right),
\end{equation}
where $u_i$ denotes the uncertainty estimate at node $i$. The inhibited local feature is
\begin{equation}
\tilde{\mathbf{l}}_i
=
\mathbf{l}_i
-
\boldsymbol{\alpha}_i
\odot
\tanh(\mathbf{l}_i-\bar{\mathbf{l}}_i).
\end{equation}
This operation encourages local consistency while retaining sharp transitions when the local evidence is reliable.

\paragraph{Uncertainty Guided Fusion.}
AdaSpX predicts an uncertainty embedding from the current, local, and surround features:
\begin{equation}
\mathbf{u}^{\mathrm{emb}}_i
=
\sigma\left(
\mathrm{MLP}
([\mathbf{x}_i;\mathbf{l}_i;\mathbf{s}_i])
\right),
\qquad
u_i=\mathrm{mean}(\mathbf{u}^{\mathrm{emb}}_i).
\end{equation}
The final fused representation is obtained by adaptively combining the inhibited local feature and surround context:
\begin{equation}
\mathbf{q}_i
=
\sigma\left(
\mathrm{MLP}
([\tilde{\mathbf{l}}_i;\mathbf{s}_i;
\tilde{\mathbf{l}}_i-\mathbf{s}_i;
\tilde{\mathbf{l}}_i\odot\mathbf{s}_i;
\mathbf{u}^{\mathrm{emb}}_i])
\right)
\odot
(1-\mathbf{u}^{\mathrm{ch}}_i),
\end{equation}
\begin{equation}
\mathbf{z}_i
=
\mathbf{q}_i\odot\tilde{\mathbf{l}}_i
+
(1-\mathbf{q}_i)\odot\mathbf{s}_i .
\end{equation}
When the predicted uncertainty is high, the fusion shifts toward broader surround context; when uncertainty is low, the block preserves stronger local evidence. The AdaSpX block is then wrapped in a standard residual FFN structure:
\begin{equation}
\mathbf{x}' = \mathbf{x} + \mathrm{DropPath}(\mathbf{W}_{o}\mathbf{z}),
\qquad
\mathbf{x}'' = \mathbf{x}' + \mathrm{DropPath}(\mathrm{FFN}(\mathrm{LN}(\mathbf{x}'))).
\end{equation}

\subsection{Acuity Lateral Lightweight Architecture}
\label{app:lite_architecture}

AdapToPASS-Swift retains the bifocal design of AdapToPASS through an \textit{Acuity stream} and a \textit{Lateral stream}. The Acuity stream emphasizes high resolution local structure, while the Lateral stream operates over coarser spherical resolutions to capture broad contextual information. Both streams are implemented using AdaSpX blocks, enabling efficient spherical aggregation without attention.

\paragraph{Acuity Stream.}
The Acuity stream processes features at the finest available icosphere resolution and uses compact neighborhoods to preserve local geometry. In our implementation, the Acuity stream uses smaller near and far neighborhoods:
\begin{equation}
K^{\mathrm{acuity}}_{\mathrm{near}}=[6,6,6],
\qquad
K^{\mathrm{acuity}}_{\mathrm{far}}=[12,8,6].
\end{equation}
This stream is responsible for maintaining fine grained spatial detail and stable local predictions.

\paragraph{Lateral Stream.}
The Lateral stream starts from a coarser spherical resolution and uses larger neighborhoods to provide wide field contextual support:
\begin{equation}
K^{\mathrm{lateral}}_{\mathrm{near}}=[12,10,8],
\qquad
K^{\mathrm{lateral}}_{\mathrm{far}}=[24,18,12].
\end{equation}
This stream captures semantic context over a broader field of view and supports the Acuity stream in ambiguous regions.

\paragraph{Uncertainty Gated Cross Stream Exchange.}
At each hierarchical level, the Acuity and Lateral streams exchange information through uncertainty gated bidirectional fusion. Let $\mathbf{x}^{l}_{a}$ and $\mathbf{x}^{l}_{b}$ denote the Acuity and Lateral features at level $l$. The module first computes cross stream messages:
\begin{equation}
\mathbf{m}_{b\rightarrow a}^{l}
=
\mathbf{W}_{b\rightarrow a}\mathrm{LN}(\mathbf{x}^{l}_{b}),
\qquad
\mathbf{m}_{a\rightarrow b}^{l}
=
\mathbf{W}_{a\rightarrow b}\mathrm{LN}(\mathbf{x}^{l}_{a}).
\end{equation}
A fusion uncertainty signal is predicted from both streams:
\begin{equation}
u_{\mathrm{fus}}^{l}
=
\sigma\left(
\mathrm{MLP}
([\mathrm{LN}(\mathbf{x}^{l}_{a});
\mathrm{LN}(\mathbf{x}^{l}_{b})])
\right).
\end{equation}
The gated updates are then given by
\begin{equation}
\mathbf{x}^{l,+}_{a}
=
\mathbf{x}^{l}_{a}
+
\mathbf{g}_{b\rightarrow a}^{l}
\odot
\mathbf{m}_{b\rightarrow a}^{l},
\end{equation}
\begin{equation}
\mathbf{x}^{l,+}_{b}
=
\mathbf{x}^{l}_{b}
+
\mathbf{g}_{a\rightarrow b}^{l}
\odot
\mathbf{m}_{a\rightarrow b}^{l}.
\end{equation}
This fusion allows fine resolution features and wide field contextual features to interact throughout the network. Under higher uncertainty, the model can rely more strongly on the Lateral stream for contextual stabilization, while confident regions preserve stronger Acuity stream detail.

\paragraph{Prediction Head.}
The final segmentation logits are predicted from the fused Acuity features:
\begin{equation}
\hat{\mathbf{Y}}
=
h_{\mathrm{seg}}(\mathbf{x}_{a}^{\mathrm{out}}),
\end{equation}
where $h_{\mathrm{seg}}$ is a lightweight segmentation head.

\subsection{Training Objective}
\label{app:lite_objective}

AdapToPASS-Swift is trained using the semantic segmentation objective
\begin{equation}
\mathcal{L}_{\mathrm{Lite}}
=
w_{\mathrm{CE}}\mathcal{L}_{\mathrm{CE}}
+
w_{\mathrm{Dice}}\mathcal{L}_{\mathrm{Dice}},
\end{equation}
where the Dice term is optional and is set to zero by default. This objective keeps the training pipeline simple while allowing the model to learn ambiguity aware spherical representations through the AdaSpX forward computation.

\begin{center}
\vspace{-6pt}
\colorbox{orange!10}{
\begin{minipage}{0.96\linewidth}
    \textbf{Takeaway.} 
    AdapToPASS-Swift provides an efficient member of the AdapToPASS family for lightweight panoramic semantic segmentation. By combining adaptive spherical context aggregation, geodesic edge conditioned local reasoning, surround context stabilization, lateral inhibition, and uncertainty gated Acuity Lateral fusion, it preserves ambiguity aware spherical perception while reducing the complexity of the full model.
\end{minipage}
}
\end{center}

\paragraph{Hyperparameter Details for AdapToPASS-Swift.}
AdapToPASS-Swift operates on rank-5 icosphere vertices with RGB inputs and an ERP grid width of 1024. It uses embedding dimension 24, dimension multipliers $[1,2,4]$, encoder depths $[2,2,2]$, bottleneck depth 2, decoder depths $[2,2]$, attention heads $[2,4,8]$, MLP ratio 4.0, dropout 0.1, and drop-path rate 0.1. \\
We train AdapToPASS-Swift for 400 epochs using AdamW with learning rate $2\times10^{-4}$, weight decay $10^{-2}$, betas $(0.9,0.999)$, cosine decay, minimum learning rate $10^{-6}$, and 5 warmup epochs. Mixed precision is enabled, gradient clipping is set to 1.0, and cross-entropy is used as the training loss. The training batch sizes is 4. RGB inputs are normalized with mean 0.5 and standard deviation 0.225, and no data augmentation is used.

\begin{table*}[t]
\centering
\small
\resizebox{\textwidth}{!}{%
\begin{tabular}{l 
>{\columncolor{green!6}}c >{\columncolor{green!6}}c >{\columncolor{green!6}}c >{\columncolor{green!6}}c >{\columncolor{green!6}}c
>{\columncolor{orange!8}}c >{\columncolor{orange!8}}c >{\columncolor{orange!8}}c >{\columncolor{orange!8}}c >{\columncolor{orange!8}}c
>{\columncolor{red!6}}c >{\columncolor{red!6}}c >{\columncolor{red!6}}c >{\columncolor{red!6}}c >{\columncolor{red!6}}c}
\toprule
\multirow{2}{*}{\textbf{Method}} 
& \multicolumn{5}{c}{\cellcolor{green!15}\textbf{Mild}} 
& \multicolumn{5}{c}{\cellcolor{orange!18}\textbf{Robust}} 
& \multicolumn{5}{c}{\cellcolor{red!15}\textbf{Stress}} \\
\cmidrule(lr){2-6} \cmidrule(lr){7-11} \cmidrule(lr){12-16}
& \textbf{Rot.} & \textbf{Scale} & \textbf{\shortstack{Ori.}} & \textbf{Trans.} & \textbf{\shortstack{View.}}
& \textbf{Rot.} & \textbf{Scale} & \textbf{\shortstack{Ori.}} & \textbf{Trans.} & \textbf{\shortstack{View.}}
& \textbf{Rot.} & \textbf{Scale} & \textbf{\shortstack{Ori.}} & \textbf{Trans.} & \textbf{\shortstack{View.}} \\
\midrule
HexRUNet       
& 37.84 & 39.01 & 35.91 & 38.29 & 34.01
& 28.94 & 34.62 & 27.53 & 33.39 & 24.60
& 19.98 & 31.06 & 18.45 & 28.67 & 15.68 \\

SphereUFormer  
& 37.40 & 53.00 & 27.10 & 35.50 & 19.80
& 30.40 & 27.20 & 23.40 & 50.70 & 19.80
& 26.40 & 38.10 & 30.30 & 13.30 & 19.10 \\

Trans4PASS     
& 57.40 & 56.83 & 54.42 & 53.78 & 54.08
& 40.10 & 50.49 & 38.63 & 45.94 & 34.12
& 25.42 & 44.31 & 25.66 & 38.44 & 18.87 \\

\midrule
\rowcolor{gray!12}
\textbf{AdapToPASS-Swift (Ours)}   
& \textbf{62.20} & \textbf{61.80} & \textbf{57.80} & \textbf{60.10} & \textbf{55.20}
& \textbf{47.80} & \textbf{54.00} & \textbf{43.80} & \textbf{50.50} & \textbf{38.90}
& \textbf{31.80} & \textbf{47.00} & \textbf{28.60} & \textbf{41.70} & \textbf{25.30} \\
\bottomrule
\end{tabular}%
}
\caption{\textbf{mIoU comparison of lightweight panoramic semantic segmentation models under unseen spherical transformations on the Stanford2D3D benchmark.} Results are reported across mild, robust, and stress settings, including rotation (Rot.), Scale, Orientation Shifts (Ori.), Translation (Trans.), and Viewpoint Shifts (View.). \textit{Higher is better.}}
\label{tab:lite_robustness_results}
\end{table*}



\end{document}